\documentclass{article}

\usepackage{microtype}
\usepackage{graphicx}
\usepackage{subcaption}
\usepackage{booktabs} 
\usepackage{tabularx}
\usepackage[most]{tcolorbox}
\usepackage{xcolor}   
\usepackage{amsmath, amssymb}

\usepackage{hyperref}
\usepackage{multirow}
\usepackage[table]{xcolor}

\newcommand{\duo}[1]{{\color{black}#1}}
\newcommand{\klj}[1]{{\color{black}#1}}
\definecolor{mygreen}{rgb}{0.0, 0.6, 0.0} 
\newcommand{\zm}[1]{\textcolor{black}{#1}}
\newcommand{\zmtwo}[1]{\textcolor{black}{#1}}

\definecolor{myback}{HTML}{DBE6E8}    
\definecolor{myframe}{HTML}{495365}   

\usepackage[preprint]{icml2026}

\usepackage{amsmath}
\usepackage{amssymb}
\usepackage{mathtools}
\usepackage{amsthm}
\usepackage{mathrsfs}

\usepackage[capitalize,noabbrev]{cleveref}

\theoremstyle{plain}

\theoremstyle{definition}

\theoremstyle{remark}

\usepackage[textsize=tiny]{todonotes}

\usepackage{fontawesome5}

\icmltitlerunning{Learning with Challenges: Adaptive Difficulty-Aware Data Generation for Mobile GUI Agent Training}

\begin{document}

\twocolumn[
  \icmltitle{\raisebox{-0.2\height}{\includegraphics[height=1.4em]{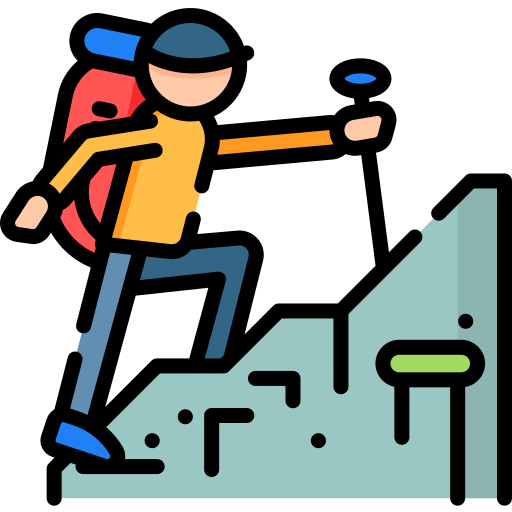}}  Learning with Challenges: Adaptive Difficulty-Aware Data Generation
  \\
  for Mobile GUI Agent Training}



  \icmlsetsymbol{equal}{*}

  \begin{icmlauthorlist}
    \icmlauthor{Linjia Kang}{thu,equal}
    \icmlauthor{Zhimin Wang}{thu,equal}
    \icmlauthor{Yongkang Zhang}{hzau,equal}
    \icmlauthor{Duo Wu}{thu}
    \icmlauthor{Jinghe Wang}{thu}
    \icmlauthor{Ming Ma}{ks}
    \icmlauthor{Haopeng Yan}{ks}
    \icmlauthor{Zhi Wang}{thu}
  \end{icmlauthorlist}

  \icmlaffiliation{thu}{Tsinghua University}
  \icmlaffiliation{hzau}{Huazhong Agricultural University}
  \icmlaffiliation{ks}{Kuaishou Technology}

  \icmlcorrespondingauthor{Zhi Wang}{wangzhi@sz.tsinghua.edu.cn}

  \icmlkeywords{Data Generation, GUI Agent, Vision Language Models, Multi-Agent}

  \vskip 0.3in
]



\printAffiliationsAndNotice{* Equal Contribution. Yongkang's work was done during his internship at Tsinghua University.}  

\begin{abstract}
\klj{Large-scale, high-quality interaction trajectories are essential for advancing mobile Graphical User Interface (GUI) agents.}
\klj{While existing methods typically rely on labor-intensive human demonstrations or automated model exploration to generate GUI trajectories, they lack fine-grained control over task difficulty. This fundamentally restricts learning effectiveness due to the mismatch between the training difficulty and the agent’s capabilities.}
\duo{Inspired by how humans acquire skills through progressively challenging tasks}, we propose \texttt{MobileGen}, a novel data generation framework that adaptively aligns training difficulty with the GUI agent’s capability frontier. 
\duo{Specifically, \texttt{MobileGen} explicitly decouples task difficulty into structural (e.g., trajectory length) and semantic (e.g., task goal) dimensions. It then iteratively evaluates the agent on a curated prior dataset to construct a systematic profile of its capability frontier across these two dimensions. With this profile, the probability distribution of task difficulty is adaptively computed, from which the target difficulty for the next round of training can be sampled. Guided by the sampled difficulty, a multi-agent controllable generator is finally used to synthesize high-quality interaction trajectories along with corresponding task instructions.}
Extensive experiments show that \texttt{MobileGen} consistently outperforms existing data generation methods by  improving the average performance of GUI agents by 1.57$\times$  across multiple challenging benchmarks. This highlights the importance of capability-aligned data generation for effective mobile GUI agent training.
\end{abstract}

\section{Introduction}
\label{introduction}

\duo{Recent advances} in Vision-Language Models (VLMs) have unlocked the potential for autonomous agents to perceive, reason, and interact directly with complex environments~\cite{gao2024generalist,zhang2024large}.
\duo{Among them, Graphical User Interface (GUI) agents~\cite{wang2025ui,ye2025mobile} establish a new paradigm for intelligent digital automation by imitating human operations to navigate mobile devices and accomplish real-world tasks~\cite{liu2025llm}.}
\duo{While early efforts often rely on prompt engineering to elicit desired behaviors from GUI agents~\cite{wen2024autodroid,wang2024mobile}, \klj{these approaches are inherently limited by inference-time reasoning, where the failed interactions cannot be internalized into model capability improvements.} Consequently, there is a growing emphasis on training agents on \klj{expert interaction trajectories} to master long-horizon, cross-application execution~\cite{liu2024autoglm,wang2025ui,ye2025mobile,tang2025magicgui,li2025coloragent}. Hence, this highlights the critical need for large-scale, high-quality, multi-step trajectory datasets that accurately capture the mapping from GUI observations to precise interaction actions.}

\duo{Unfortunately, constructing high-quality training datasets remains a primary bottleneck in boosting the performance of GUI agents~\cite{liu2024autoglm,zhang2024large}. Conventional approaches rely on human \klj{demonstrations}~\cite{lu2025guiodyssey,zhang2025tongui,jang2025scalable}, \klj{which typically require either expensive manual annotations or access to restricted online resources. Recent model-based methods leverage automated strategies, such as VLMs or random walks}, to autonomously generate trajectory data through environment exploration~\cite{lin2025gui,xiao2025ui,xu2025agenttrek,sun2025genesis}, with better scalability and \klj{lower cost of data collection}. \klj{Nevertheless, these methods still suffer from a key limitation: the absence of explicit mechanisms for modeling and controlling trajectory difficulty during data generation.} As a result, they cannot produce trajectories whose difficulty dynamically aligns with the capabilities of agents, ultimately hindering effective learning~\cite{chen2025agentfrontier,yang2025zerogui}.}

\duo{Much like humans acquire complex skills by progressively engaging with more challenging tasks~\cite{guadagnoli2004challenge}, \klj{GUI agents tend to} learn most effectively when trained on trajectories whose difficulty is well-matched to their current capabilities. For instance, a simple trajectory (e.g., clicking an obvious  button) provides little learning signal for improvement, whereas an overly complex one (e.g., \klj{navigating through multiple nested menus}) may overwhelm the agent and impede learning. This suggests that the optimal challenge point for agent training lies at the intersection of data difficulty and the agent’s capability: a regime that is neither so simple as to be uninformative nor so complex as to be intractable.}

\duo{Drawing inspiration from the above insight, we propose \texttt{MobileGen}, an adaptive data generation framework for training GUI agents. By dynamically setting the challenge point that align data difficulty with the agent's current capability, \texttt{MobileGen} continuously pushes the frontier of the agent's capability \klj{while maximizing training effectiveness.}}
\klj{Specifically, \texttt{MobileGen} first decouples GUI trajectory difficulty into \textit{structural difficulty} and \textit{semantic difficulty}, providing the basis for fine-grained control. Then the generation process is organized into three stages. First, in \textit{agent capability profiling} stage, we infer the agent’s dual-level capability frontier by evaluating its performance on our curated prior dataset. Next, in \textit{difficulty distribution generation} stage, we build adaptive task difficulty distributions over the joint structural and semantic space with $\alpha$-guided challenge point, and drive the generation toward challenges aligned with the agent’s current capabilities. Finally, in \textit{difficulty-aware trajectory generation} stage, parameters sampled from these distributions drive our Multi-agent Controllable Generator (MCG) to efficiently produce high-quality interaction trajectories in parallel. Leveraging distribution-level difficulty control, \texttt{MobileGen} ensures stable, scalable, and customized data generation, significantly improving training effectiveness for GUI agents. 
We summarize our contributions as follows:}
\klj{
\begin{itemize}
    \item We propose \texttt{MobileGen}, a capability-aligned data generation framework that explicitly matches training difficulty to an agent’s capability frontier, establishing a new paradigm for effective GUI agent training.
    \item We design a distribution-level difficulty control mechanism for GUI trajectory generation by decoupling difficulty into structural and semantic dimensions, enabling controllable and efficient synthesis of high-quality interaction trajectories.
    \item  Extensive experiments on multiple  benchmarks show that \texttt{MobileGen} consistently generates high-quality, challenging training trajectories, improving performance by nearly 1.57$\times$ over zero-shot baselines, and consistently outperforming existing SOTA methods.
\end{itemize}

}


\begin{figure*}[t]
    \centering
    \includegraphics[width=0.98\linewidth]{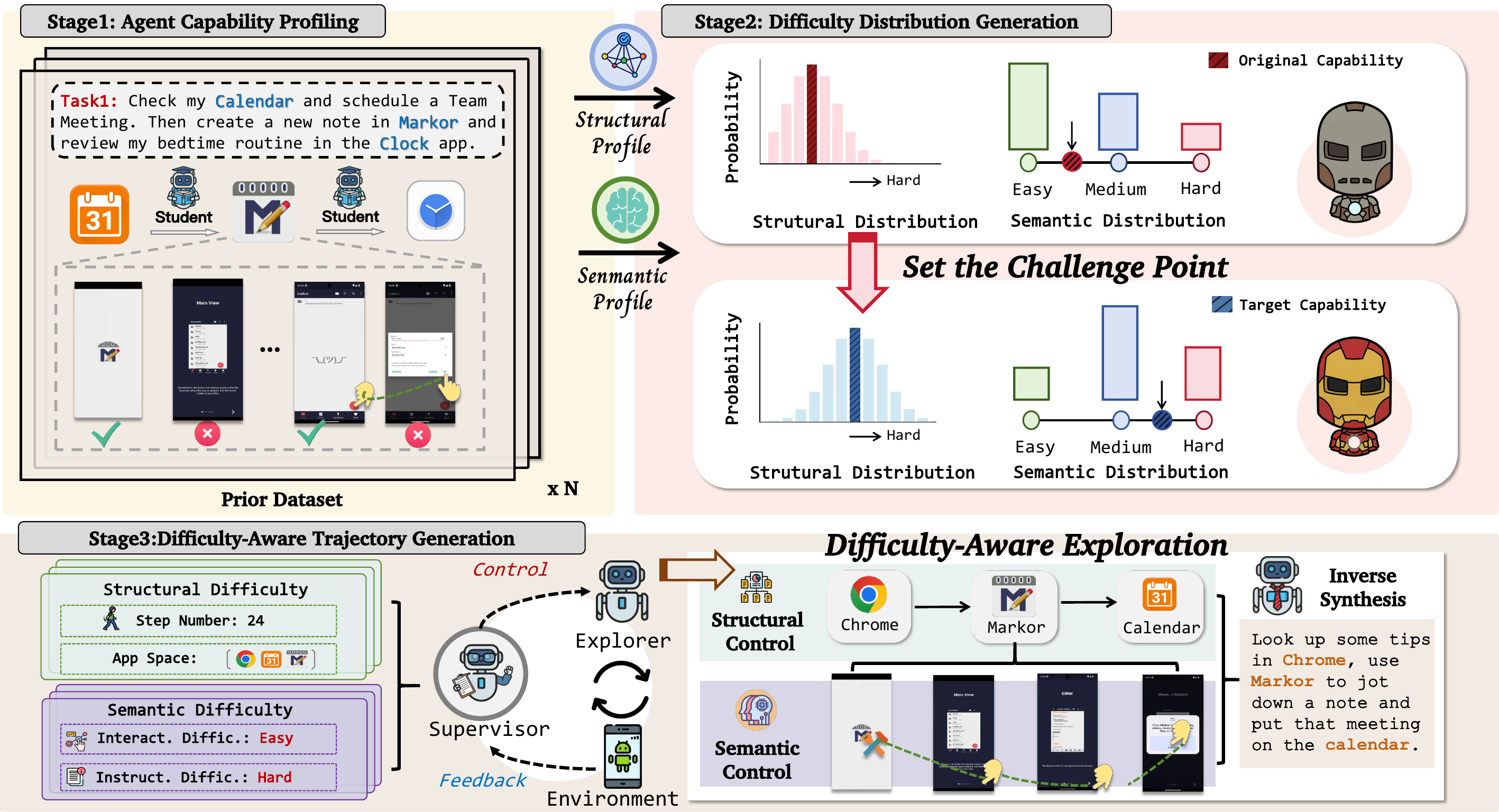}
    \caption{\zm{\klj{Overview of \texttt{MobileGen}. Our pipeline consists of three key stages:} (1) \textit{Agent Capability Profiling}: \klj{The student agent $\mathcal{M}_\text{student}$} is evaluated on a prior dataset to derive structural and semantic capability profiles. 
    (2) \textit{Difficulty Distribution Generation}: Based on the profiles, the challenge point is set to form the desired difficulty distribution. 
    (3) \textit{Difficulty-Aware Trajectory Generation}: Guided by the sampled difficulty parameters, \klj{two agents $\mathcal{M}_\text{explorer}$ and $\mathcal{M}_\text{supervisor}$} collaborate to generate \klj{interaction trajectories}. The reasoning traces and instructions will then be reconstructed \klj{via inverse synthesis.}}}
    \label{fig:first-figure}
\end{figure*}

\section{Related Work}
\subsection{Graphical User Interface Agents}
\zm{
Recent advances in VLMs~\cite{bai2025qwen2,wang2025internvl3} have significantly accelerated the development of autonomous GUI agents. Current approaches for GUI agent design generally fall into two main categories~\cite{liu2025llm,hu2025agents}: prompt-based methods and training-based methods. Prompt-based methods~\cite{wen2024autodroid,wang2024mobile,wen2023droidbot} 
 are lightweight and easy to deploy, but often exhibit limited scalability and struggle with complex tasks due to the lack of domain-specific knowledge for GUI operations. Consequently, training-based methods~\cite{ye2025mobile,wang2025ui,cheng2024seeclick,chen2025guicourse,zhang2025ufo2} 
 adapt VLMs via Supervised Fine-Tuning (SFT)~\cite{ye2025mobile,liu2024autoglm} or Reinforcement Learning (RL)~\cite{wang2025ui,li2025efficient,tang2025magicgui}, 
with SFT being widely adopted due to its simplicity and stability. Nevertheless, the performance of training-based approaches remains fundamentally constrained 
by the scarcity of high-quality human interaction trajectories.}

\subsection{Data Generation for GUI Agents}

\zm{While text-based agents have demonstrated significant progress in structured reasoning and planning~\cite{schick2023toolformer,jin2025search,wu2025catp,wang2025cobel}, GUI agents face unique challenges due to the vast, unstructured visual state space and thus require large-scale, high-quality training data.}
Early efforts primarily rely on human demonstrations~\cite{rawles2023androidinthewild,li2024effects,lu2025guiodyssey}, \zm{which are labor-intensive and inherently difficult to scale.}
\zm{Although some approaches~\cite{jang2025scalable,zhang2025tongui} attempt to mitigate this by transforming tutorial videos into interaction trajectories, they remain constrained by the limited availability and coverage of online instructional content.}

\zm{To address these scalability bottlenecks, recent works have explored model-based methods~\cite{lin2025gui,sun2025genesis,xie2025agentsynth,yang2025zerogui,xiao2025ui}, leveraging VLMs to autonomously explore GUI environments and generate interactions that mimic human operations. For instance, OS-Genesis~\cite{sun2025genesis} proposes a reverse task synthesis framework to generate instructions from stochastic exploration trajectories}, while GUI-ReWalk~\cite{lin2025gui} refines stochastic exploration through intent-aware reasoning. \zm{AgentSynth~\cite{xie2025agentsynth} chains executable subtasks into complex, long-horizon datasets by exploiting information asymmetry. To ensure data quality, \citet{xiao2025ui} further integrates a learned reward model for trajectory filtering.} \zm{However, current model-based methods lack explicit difficulty control and thus often generate data misaligned with the agent capability, \klj{leading to learning degradation}. To bridge this gap, \klj{we propose \texttt{MobileGen} to dynamically align training difficulty with the agent proficiency, thereby guiding training toward a suitable challenge point.}}

\section{Methodology}
\subsection{Preliminaries}\label{sec:preliminaries}
Aligning training task difficulty with an agent’s capability requires fine-grained modeling of GUI task trajectories. For GUI tasks, a trajectory can be formulated as a goal-conditioned sequence of state-action tuples, $\tau = \{g, (s_1, a_1), \ldots, (s_n, a_n)\}$, where $g$ is the trajectory-level task goal, each state $s_t$ includes multimodal observations (e.g., screenshots), and each action $a_t$ corresponds to low-level GUI manipulations (e.g., click). The difficulty of a trajectory is determined not only by its structural properties (e.g., trajectory length), but also by semantic factors such as task goal specification and GUI state transitions under this formulation. Based on the insight, we decompose trajectory difficulty into two complementary dimensions:

\noindent \textbf{Structural Difficulty.} This captures the decision-making challenges arising from a trajectory's interaction structure, quantified using two metrics: (1) \emph{Depth of Trajectory (DoT)}, the total number of steps within a trajectory, reflecting the need for long-horizon reasoning. (2) \emph{Breadth of Trajectory (BoT)}, the number of distinct applications within a trajectory, reflecting the complexity of maintaining context across applications. Both DoT and BoT are positive integers.

\noindent \textbf{Semantic Difficulty.} This represents the cognitive challenges involved in understanding and executing task goals, assessed along two metrics: (1) \emph{Interaction Control Difficulty (ICD)}, the difficulty of mapping goals to actions across interfaces. (2) \emph{Instruction Understanding Difficulty (IUD)}, the difficulty of interpreting natural language instructions and inferring implicit user intent. Given the inherent ambiguity, both ICD and IUD are discretized into $\{easy, medium, hard\}$.

Altogether, these four metrics jointly serve as parameters for trajectory difficulty and will be employed in Section~\ref{sec:mcg} to generate trajectories that satisfy specific structural and semantic difficulty constraints, thereby enabling fine-grained control over the difficulty of training tasks.

\subsection{\texttt{MobileGen}}

In this section, we present the overall pipeline of \texttt{MobileGen}, which covers the process from customizing the training task distribution to controllable generation of high-quality trajectories. \zm{As shown in Figure \ref{fig:first-figure}}, our pipeline is organized into three stages: (1) \textit{agent capability profiling}, (2) \textit{difficulty
distribution generation} and (3) \textit{difficulty-aware trajectory generation}, as detailed below.

\subsubsection{Agent Capability Profiling}\label{sec:profiling}
We first conduct a systematic assessment of the current capability of the student agent $\mathcal{M}_{\text{student}}$, defined as the agent to be tuned, based on a carefully designed prior dataset $\mathscr{T}_p$ covering trajectories of varying difficulty. We then adopt a dual-level assessment paradigm that jointly incorporates \textit{structural profiling} and \textit{semantic profiling} in agent capability modeling. Evaluation on $\mathscr{T}_p$ yields capability profiles of $\mathcal{M}_{\text{student}}$, which serves as a unified reference for guiding difficulty regulation in subsequent training task generation.

\noindent \textbf{Structural Profiling.}
To characterize the competence of $\mathcal{M}_{\text{student}}$ in executing GUI tasks of varying structural difficulty, 
we derive three statistics from the prior dataset $\mathscr{T}_p$.
Specifically, sequential execution capability $\mathcal{C}_{d}$ approximates the length of task trajectories that $\mathcal{M}_{\text{student}}$ can reliably execute, and is used to guide the distribution of DoT in training tasks.
Cross-application interaction capability $\mathcal{C}_{b}$ measures the average number of distinct applications that the model can reliably interact within a trajectory, informing the distribution design of BoT.
In addition, application vulnerability score $\mathcal{V}_i$ quantifies the relative failure frequency of the model on application $\text{app}_i$, which enables the identification of application-specific weak areas.
These metrics are formally defined in Equation~\eqref{structural_profile}.
\begin{equation}\label{structural_profile}
\left\{
\begin{aligned}
\mathcal{C}_{d} &=
\frac{1}{|\mathscr{T}_p|}
\sum_{\tau_p \in \mathscr{T}_p}
\sum_{t \in \tau_p}
\mathbb{I}_{\{\hat{a}_t = a_t\}}, \\
\mathcal{C}_{b} &=
\frac{1}{|\mathscr{T}_p|} \sum_{\tau_p \in \mathscr{T}_p} \sum_{\text{app} \in \mathcal{A}_{\tau_p}} \frac{\sum_{t \in \tau_p} \mathbb{I}_{\{\text{app}_t = \text{app} \wedge \hat{a}_t = a_t\}}}{\sum_{t \in \tau_p} \mathbb{I}_{\{\text{app}_t = \text{app}\}}}\\
\mathcal{V}_i &=
\frac{
\sum_{\tau_p \in \mathscr{T}_p}
\sum_{t \in \tau_p}
\mathbb{I}_{\{\text{app}_t = \text{app}_i \land \hat{a}_t \neq a_t\}}
}{
\sum_{\tau_p \in \mathscr{T}_p}
\sum_{t \in \tau_p}
\mathbb{I}_{\{\text{app}_t = \text{app}_i\}}
}.
\end{aligned}
\right.
\end{equation}
Here, $\mathbb{I}_{\{\cdot\}}$ is the indicator function, equal to $1$ if the condition holds and $0$ otherwise. $\hat{a}^{(t)}$ and $a^{(t)}$ denote the predicted and ground-truth actions at step $t$, respectively. $\text{app}_{t}$ denotes the application associated with step $t$. For a trajectory $\tau_p$, $\mathcal{A}_{\tau_p}$ denotes the application set visited in $\tau_p$.

\noindent \textbf{Semantic Profiling.}
We further assess the competence of $\mathcal{M}_{\text{student}}$ in handling GUI tasks of varying semantic difficulty. We introduce two semantic capability metrics:
interaction control capability $\mathcal{C}_{\text{int}}$ that reflects the model’s ability to translate task intent into correct interaction behaviors,
and instruction understanding capability $\mathcal{C}_{\text{ins}}$ that captures its ability to comprehend and reason over natural language instructions.
Both metrics are formalized in Equation~\eqref{semantic_profile}.
Here, $m^{(t)}_{\text{interact}}$ and $m^{(t)}_{\text{instruct}}$ are the preprocessed semantic difficulty score for step $t$, corresponding to interaction control and instruction understanding, respectively.
\begin{equation}\label{semantic_profile}
\left\{
\begin{aligned}
\mathcal{C}_{\text{int}} &=
\frac{
\sum_{\tau_p \in \mathscr{T}_p}
\sum_{t \in \tau_p}
\mathbb{I}_{\{\hat{a}_t = a_t\}} * m^{(t)}_{\text{int}}
}{
\sum_{\tau_p \in \mathscr{T}_p}
\sum_{t \in \tau_p} 
\mathbb{I}_{\{\hat{a}_t = a_t\}}
}, \\
\mathcal{C}_{\text{ins}} &=
\frac{
\sum_{\tau_p \in \mathscr{T}_p}
\sum_{t \in \tau_p}
\mathbb{I}_{\{\hat{a}_t = a_t\}} * m^{(t)}_{\text{ins}}
}{
\sum_{\tau_p \in \mathscr{T}_p}
\sum_{t \in \tau_p}
\mathbb{I}_{\{\hat{a}_t = a_t\}}
}.
\end{aligned}
\right.
\end{equation}

We adopt Pass@K~\cite{chen2021pass_at_k} to evaluate whether the predicted actions match the ground-truth actions in $\mathscr{T}_p$ for fully exploiting $\mathcal{M}_{\text{student}}$’s potential for improvement. For each step, $\mathcal{M}_{\text{student}}$ takes as input the SoM-annotated~\cite{yang2023som} screenshots along with the corresponding UI-Tree to guide its action prediction. More details of $\mathscr{T}_p$ are provided in Appendix~\ref{apdx:prior}.

\begin{figure*}[t]
    \centering
    \includegraphics[width=0.98\linewidth]{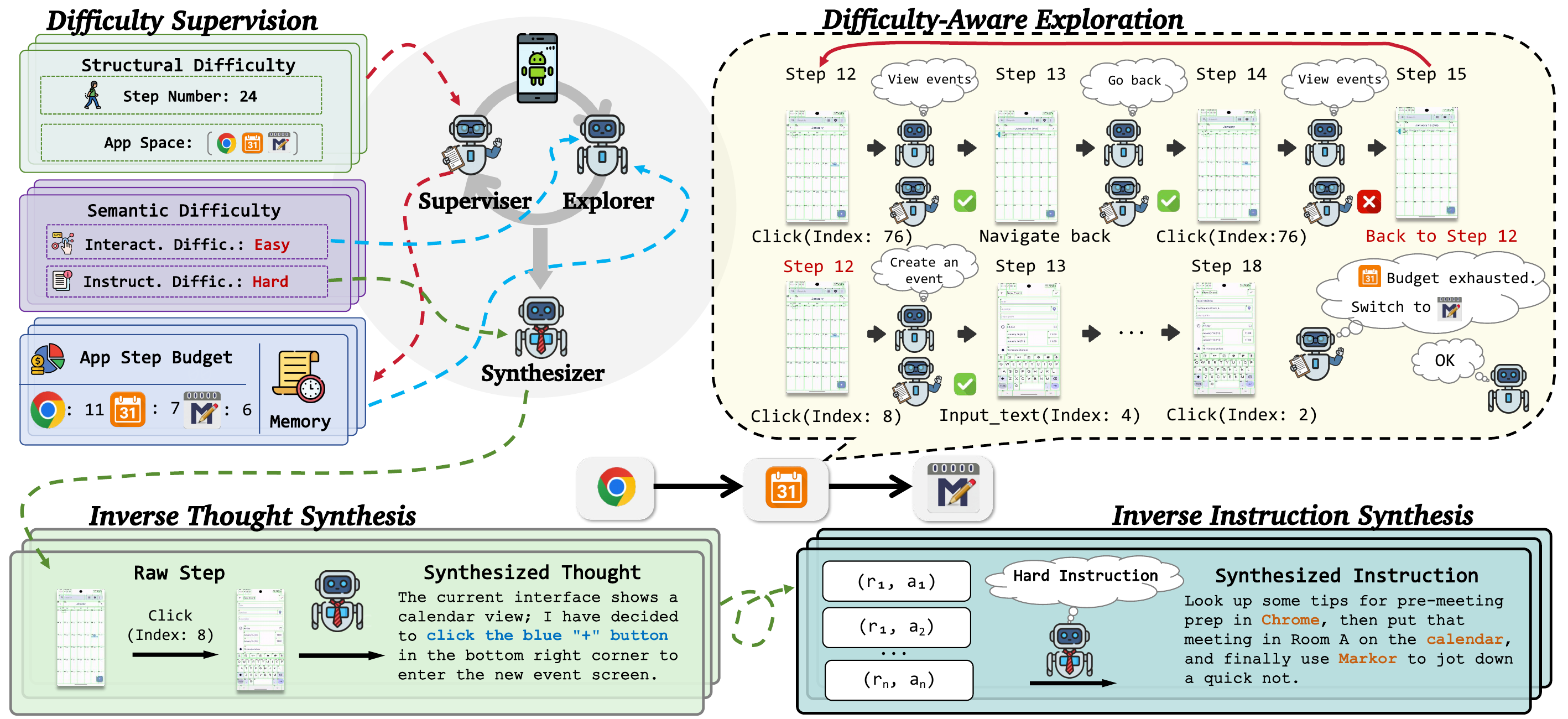}
    \caption{\klj{Detailed workflow of MCG. Concretely, MCG generates difficulty-aware training trajectories controlled by sampled difficulty parameters through multi-agent collaboration.
    During \textit{Interaction Trajectory Generation}, the supervisor $\mathcal{M}_{\text{supervisor}}$ allocates exploration step budgets for each application before the exploration begins and dynamically manages the explorer $\mathcal{M}_{\text{explorer}}$’s step usage across applications while performing rollbacks to correct interaction errors. 
    Following exploration, the synthesizer $\mathcal{M}_{\text{synthesizer}}$ reconstructs step-level thoughts and trajectory-level instructions.
    }}
    \label{fig:second-figure}
\end{figure*}

\subsubsection{Difficulty Distribution Generation}\label{sec:distribution_generation}
In this part, we transform the capability profile of $\mathcal{M}_{\text{student}}$ into a set of task sampling distributions that explicitly control the training task difficulty.
In particular, a challenge point $\alpha$ is introduced to control the challenge of training tasks, and the evolution objectives that it guides are applied independently to generate the structural and semantic difficulty distributions.

\noindent\textbf{$\alpha$-Guided Challenge Point.}
Given the capability statistics calculated with Equation~\eqref{structural_profile} and~\eqref{semantic_profile} during profiling, we define the challenge point for each capability dimension.
For a generic capability measure $\mathcal{C}$, the target value is computed as Equation~\eqref{eq:alpha}.
\begin{equation}\label{eq:alpha}
\mathcal{C}^{\star} = \mathcal{C}\,\bigl(1 + \alpha*\eta\bigr),
\end{equation}
where $\alpha > 0$ controls the overall aggressiveness of the training tasks and $\eta > 0$ is a pre-defined, dimension-specific constant that characterizes the relative rate of evolution for different capability dimensions.
By adjusting $\alpha$, the training tasks can be tuned to shift toward more challenging tasks, thereby explicitly defining the training challenge point.

\noindent\textbf{Structural Difficulty Distribution.}\label{structural_distribution}
For task structural difficulty parameters, we construct discrete sampling distributions centered at their corresponding challenge points. Specifically, let $x$ denote a candidate value for the structural difficulty parameters (i.e., DoT and BoT for the training tasks). Given a target structural capability $\mathcal{C}^{\star}$,
we define a finite sampling range and assign probabilities to each $x$ using gaussian kernel function,
\begin{equation}\label{eq:gaussian}
p_\text{structural}(x) = \frac{\exp\!\left(-\frac{(x - \mathcal{C}^{\star})^2}{2\sigma^2}\right)}{\sum_x \exp\!\left(-\frac{(x - \mathcal{C}^{\star})^2}{2\sigma^2}\right)},
\end{equation}
where $\sigma$ is a hyperparameter controlling the spread of the distribution. The use of the gaussian kernel is motivated by its property of smoothly concentrating probability mass around the challenge point, thereby biasing sampling toward tasks whose structural difficulty closely matches $\mathcal{M}_{\text{student}}$’s current learning frontier.

\noindent\textbf{Semantic Difficulty Distribution.}
Similarly, let the target semantic capability be \(\mathcal{C}^{\star}\), we define the affinity of \(\mathcal{C}^{\star}\) to each semantic difficulty level under Equation~\eqref{membership},

\begin{equation}\label{membership}
\left\{
\begin{aligned}
\mu_\text{easy}(\mathcal{C}^{\star}) &= \max\Big(0, \min\big(1, \frac{c_\text{mid} - \mathcal{C}^{\star}}{c_\text{mid} - c_\text{min}}\big)\Big), \\
\mu_\text{medium}(\mathcal{C}^{\star}) &= \max\Big(0, \min\big(\frac{\mathcal{C}^{\star} - c_\text{min}}{c_\text{mid} - c_\text{min}}, \frac{c_\text{max} - \mathcal{C}^{\star}}{c_\text{max} - c_\text{mid}}\big)\Big), \\
\mu_\text{hard}(\mathcal{C}^{\star}) &= \max\Big(0, \min\big(1, \frac{\mathcal{C}^{\star} - c_\text{mid}}{c_\text{max} - c_\text{mid}}\big)\Big),
\end{aligned}
\right.
\end{equation}

where \(c_\text{min}, c_\text{mid}, c_\text{max}\) correspond to the semantic difficulty levels mapped to easy, medium, and hard, respectively.
These affinities are then normalized to form a sampling distribution over semantic difficulty levels of training tasks:

\begin{equation}
p_\text{semantic}(l) = \frac{\mu_l(\mathcal{C}^{\star})}{\sum_{l \in \{\text{easy, medium, hard}\}} \mu_{l}(\mathcal{C}^{\star})}.
\end{equation}

This distribution naturally reflects the alignment between $\mathcal{M}_{\text{student}}$’s challenge point and each difficulty level: the higher the affinity of a difficulty level, the greater its probability of being sampled, and vice versa.

Our distribution design ensures that the difficulty of training tasks both aligns with $\mathcal{M}_{\text{student}}$’s capability challenge point and allows for moderate exploration of other difficulty levels, ensuring both stability and diversity in the training tasks.

\subsubsection{Difficulty-aware Trajectory Generation}\label{sec:mcg}
Each training task consists of a set of trajectories $\mathscr{T}$. 
For each trajectory $\tau$, we sample its difficulty parameters from the structural and semantic difficulty distributions obtained in Section~\ref{sec:distribution_generation}. These parameters include DoT, BoT, ICD, and IUD, as defined in Section~\ref{sec:preliminaries}.
In particular, we sample $c_{\tau}$ applications from the application selection distribution guided by the application vulnerability scores $\mathcal{V}_i$ described in Section~\ref{sec:profiling}, forming the application space $\mathcal{A}_\tau$ for trajectory $\tau$.
These sampled parameters are then fed into our designed MCG to generate trajectories, where both structural and semantic difficulty constraints are explicitly enforced to ensure consistency with the specified difficulty parameters. \zm{As illustrated in Figure \ref{fig:second-figure}}, 
MCG consists of two sub-stages: (1) \textit{interaction trajectory generation} and (2) \textit{inverse synthesis}.

\noindent\textbf{Interaction Trajectory Generation.}
To generate primitive interaction trajectories reliably, we design two collaborative agents: an explorer $\mathcal{M}_{\text{explorer}}$ and a supervisor $\mathcal{M}_{\text{supervisor}}$.
Specifically, at each interaction step, $\mathcal{M}_{\text{explorer}}$ takes a screen screenshot annotated with Set-of-Mark (SoM)~\cite{yang2023som}, the corresponding UI tree, together with the history $\tilde{\mathcal{H}}$ so far~(summarized by $\mathcal{M}_{\text{supervisor}}$ based on the global working memory $\mathcal{H}$) as input. It is guided to perform human-like interactions in GUI environments under the constraint of the interaction control difficulty parameter ICD. 
\klj{This constraint is realized by encoding the prompt corresponding to ICD into the prompts of $\mathcal{M}_{\text{explorer}}$, which shapes its exploration strategy and interaction behavior.
Meanwhile, $\mathcal{M}_{\text{supervisor}}$ provides global regulation to maintain stable data difficulty and quality.
This regulation is realized by:
(1) A \textit{step budget} mechanism that dynamically allocates exploration steps across applications based on DoT, ensuring balanced coverage of the application space $\mathcal{A}_\tau$.
(2) A \textit{rollback} mechanism that recovers from anomalous interaction patterns within the current application according to $\mathcal{H}$, which  contributes to stabilizing trajectory quality during long-horizon exploration. 
Together, these two mechanisms enable  $\mathcal{M}_{\text{supervisor}}$ to regulate exploration effectively.}

\noindent\textbf{Inverse Synthesis.}
After collecting raw interaction trajectories, we perform inverse synthesis using a synthesizer agent $\mathcal{M}_{\text{synthesizer}}$ to reconstruct the step-level reasoning traces and trajectory-level instructions.
For the $t$-th step of a trajectory $\tau$, we construct a multimodal state transition tuple $(\tilde{s}_t, a_t, s_{t+1})$, where $\tilde{s}_t$ consists of the SoM-annotated screenshot and the corresponding UI tree.
Based on this tuple, $\mathcal{M}_{\text{synthesizer}}$ infers the intermediate thought $r_t$ underlying the executed action.
Conditioned on the trajectory-level semantic difficulty parameter IUD, $\mathcal{M}_{\text{synthesizer}}$ integrates the step-wise reasoning traces $r_t$ and actions $a_t$ to produce a trajectory-level instruction of which understanding difficulty is explicitly controlled. Concretely, the prompt corresponding to IUD is injected into the prompts of $\mathcal{M}_{\text{synthesizer}}$, allowing the model to regulate the level of ambiguity of the synthesized trajectory-level instructions.

Details of sampling procedure and the application selection distribution are provided in Appendix~\ref{apdx:sampling}, and the implementation details and prompts of MCG are provided in  Appendix~\ref{apdx:mcg}.

\definecolor{A}{HTML}{E6E8FE} 
\definecolor{B}{HTML}{FCE6D5}
\definecolor{C}{HTML}{FADBDF}


\definecolor{softyellow}{RGB}{253, 251, 237}

\definecolor{darkred}{RGB}{180, 0, 0} 
\definecolor{1}{HTML}{FFFDEF} 
\definecolor{2}{HTML}{FFF7DB} 
\definecolor{white}{RGB}{255, 255, 255} 

\newcommand{\gain}[1]{\smash{$_{\textcolor{darkred}{+#1}}$}}
\newcommand{\bgain}[1]{\smash{$_{\textcolor{darkred}{\mathbf{+#1}}}$}}

\newcommand{\hlt}[2]{%
  \bgroup
  \setlength{\fboxsep}{1.5pt} 
  \colorbox{#1}{#2}%
  \egroup
}

\begin{table*}[t]
\centering
\footnotesize
\renewcommand{\arraystretch}{0.9}
\caption{Results on AndroidWorld and AndroidControl-Curated. ``SR'' denotes task success rate, ``Type'' indicates action-type accuracy, and ``Grounding'' represents grounding accuracy computed on annotated subsets. Results in \textbf{\hlt{A}{co}\hlt{B}{lo}\hlt{C}{red}} highlights and \underline{underline} denote the best and second-best results of each trained backbone, respectively.} 
\resizebox{0.95\textwidth}{!}{
\begin{tabular}{l c ccc ccc}
\toprule
\multirow{2}{*}{\textbf{Model}}
& \textbf{AndroidWorld}
& \multicolumn{3}{c}{\textbf{AndroidControl-Curated-Hard}}
& \multicolumn{3}{c}{\textbf{AndroidControl-Curated-Easy}} \\
\cmidrule(lr){2-2} \cmidrule(lr){3-5} \cmidrule(lr){6-8}
& \textbf{SR (\%)}
& \textbf{Type (\%)} & \textbf{Grounding (\%)} & \textbf{SR (\%)}
& \textbf{Type (\%)} & \textbf{Grounding (\%)} & \textbf{SR (\%)} \\
\midrule
GPT-5 & 46.6 & 66.2 & 14.5 & 16.6 & 79.0 & 16.1 & 29.6 \\
GUI-Owl-7B & 54.3 & 62.2 & 64.5 & 41.6 & 71.0 & 83.9 & 61.8 \\
UI-TARS-1.5-7B & 39.7 & 62.6 & 37.6 & 28.2 & 70.6 & 64.9 & 51.6 \\
\midrule
\rowcolor{1} \multicolumn{8}{l}{\textbf{\textit{Zero-shot}}} \\

Qwen3-VL-4B-Inst. & 25.0 & 70.8 & 26.4 & 25.6 & 81.4 & 49.1 & 37.8 \\
Qwen3-VL-8B-Inst. & 31.0 & 71.2 & 27.9 & 24.0 & 78.6 & 51.9 & 34.2 \\
InternVL3-14B & 29.3 & 69.4 & 32.7 & 27.6 & 78.0 & 47.5 & 33.6 \\
\midrule

\rowcolor{1} \multicolumn{8}{l}{\textbf{\textit{OS-Genesis}}} \\
Qwen3-VL-4B-Inst.   & 31.9 & \underline{72.8} & \underline{29.7} & 27.8 & \underline{83.6} & \underline{68.4} & \underline{54.4} \\
Qwen3-VL-8B-Inst.   & 37.9 & \underline{75.2} & 31.2 & 29.6 & \underline{85.4} & \underline{69.6} & \underline{56.8} \\
InternVL3-14B & \underline{35.3} & 71.6 & \underline{37.9} & 31.2 & \underline{81.8} & \hlt{C}{\textbf{69.9}} & \underline{55.0} \\

\midrule

\rowcolor{1} \multicolumn{8}{l}{\textbf{\textit{GUI-Rewalk}}} \\
Qwen3-VL-4B-Inst.   & \underline{33.6} & 71.8 & 28.8 & \underline{29.2} & 82.8 & 67.1 & 53.2 \\
Qwen3-VL-8B-Inst.   & \underline{41.4} & 73.0 & \underline{33.3} & \underline{31.6} & 84.2 & 68.7 & 55.8 \\
InternVL3-14B & 31.9 & \underline{73.8} & 36.4 & \underline{32.8} & 80.6 & 66.5 & 53.6 \\

\midrule

\rowcolor{2} \multicolumn{8}{l}{\textbf{\textit{Ours}}} \\
Qwen3-VL-4B-Inst. & \hlt{A}{\textbf{40.5}} & \hlt{A}{\textbf{77.6}} & \hlt{A}{\textbf{37.9}} & \hlt{A}{\textbf{35.6}} & \hlt{A}{\textbf{86.8}} & \hlt{A}{\textbf{70.9}} & \hlt{A}{\textbf{58.6}} \\
Qwen3-VL-8B-Inst.   & \hlt{B}{\textbf{45.7}} & \hlt{B}{\textbf{80.8}} & \hlt{B}{\textbf{45.5}} & \hlt{B}{\textbf{38.4}} & \hlt{B}{\textbf{88.6}} & \hlt{B}{\textbf{72.2}} & \hlt{B}{\textbf{61.2}} \\
InternVL3-14B & \hlt{C}{\textbf{42.2}} & \hlt{C}{\textbf{78.6}} & \hlt{C}{\textbf{46.7}} & \hlt{C}{\textbf{36.6}} & \hlt{C}{\textbf{86.8}} & \underline{68.4} & \hlt{C}{\textbf{58.8}} \\

\bottomrule
\end{tabular}
}

\label{tab:androidcontrol_annotated}
\end{table*}

\section{Experiments}

\subsection{Experimental Settings}
\noindent\textbf{Benchmarks.} 
We employ three representative benchmarks as our evaluation testbed. (1) \textbf{AndroidWorld}~\cite{rawles2024androidworld}, a challenging online benchmark operating in android emulators, designed to demonstrate the practicality of agents in solving real-world tasks. (2) \textbf{AndroidControl-Curated}~\cite{leung2025androidcontrol}, a refined version of AndroidControl~\cite{li2024effects}, supporting both low-level and high-level tasks. (3) \textbf{GUIOdyssey}~\cite{lu2025guiodyssey}, a comprehensive dataset focused on cross-app and long-horizon mobile GUI tasks. \klj{We adopt the official evaluation metrics defined by each respective benchmark. Details of benchmarks are illustrated in Appendix \ref{apdx:benchmark}.}

\noindent\textbf{Baselines.} 
\klj{We compare \texttt{MobileGen} against two primary categories of baselines. First, we assess the zero-shot inference performance of: (1) General-purpose VLMs, including both closed-source frontiers (e.g., GPT-5~\cite{singh2025gpt5}) and open-source families (e.g., Qwen3-VL~\cite{bai2025qwen3VL} and InternVL3~\cite{zhu2025internvl3}); and (2) Specialized GUI Agents (e.g., GUI-Owl-7B~\cite{ye2025mobile} and UI-TARS-1.5-7B~\cite{qin2025ui}), which are specially optimized on GUI tasks. Second, we compare our approach against SOTA data synthesis frameworks, including OS-Genesis~\cite{sun2025genesis}, which generates instructions via reverse task synthesis from stochastic exploration, and GUI-Rewalk~\cite{lin2025gui}, a multi-stage framework enhanced by intent-aware reasoning. We evaluate the effectiveness of these frameworks by performing SFT on three base VLMs—Qwen3-VL-4B/8B-Instruct, and InternVL3-14B—using their respective synthesized data to compare the resulting performance gains across diverse GUI benchmarks.}

\noindent\textbf{Implementation Details.} 
\klj{To ensure a fair comparison, we employ Qwen3-VL-8B-Instruct~\cite{bai2025qwen3VL} as the backbone for data generation and standardize the data scale for training as 500 trajectories with an average length of approximately 20 steps. We perform SFT on all models with LoRA~\cite{hu2022lora} on 8$\times$NVIDIA RTX 4090 GPUs. More details can be found in Appendix~\ref{apdx:exp_details}.}


\subsection{Main Results}\label{exp:main_results}
\noindent \textbf{AndroidWorld.} To verify the effectiveness of data synthesized by \texttt{MobileGen} in dynamic GUI environments, we evaluate it on AndroidWorld~\cite{rawles2024androidworld} using the standard M3A agent framework across all backbones. As shown in Table~\ref{tab:androidcontrol_annotated}, \texttt{MobileGen} significantly narrows the performance gap between open-source general-purpose VLMs and two SOTA models, GPT-5 and GUI-Owl-7B. Notably, Qwen3-VL-8B-Instruct~\cite{bai2025qwen3VL} fine-tuned with our method achieves a performance improvement of up to 11.5\% compared to specialized GUI agent UI-TARS-1.5-7B. Furthermore, \texttt{MobileGen} achieves the best results among synthesis frameworks, with Success Rates (SR) averaging 1.51$\times$ higher than zero-shot baselines. \zmtwo{This superiority suggests that proper challenge point settings lead to more effective learning under the same data scale,  underscoring the value of difficulty-aware, high-quality data in enhancing the autonomous interaction of VLMs in online GUI environments.}

\noindent\textbf{AndroidControl-Curated.} We further conduct experiments on the AndroidControl-Curated~\cite{leung2025androidcontrol}, where only 20 out of all included applications were seen during training, making it a rigorous test for Out-of-Distribution~(OOD) generalization. While the \textit{Easy} setting provides step-by-step guidance, the \textit{Hard} setting requires autonomous planning for long-horizon tasks. As shown in Table~\ref{tab:androidcontrol_annotated}, \texttt{MobileGen} consistently improves execution and planning across all backbones. It excels particularly in the \textit{Hard} setting, achieving a peak 1.44$\times$ SR increase, outperforming other synthesis methods. \zmtwo{These results suggest that our difficulty-aware paradigm enables the model to learn more effectively around the challenge point and can effectively foster generalization to challenging OOD scenarios.}

\begin{figure}
    \centering
    \includegraphics[width=0.8\linewidth]{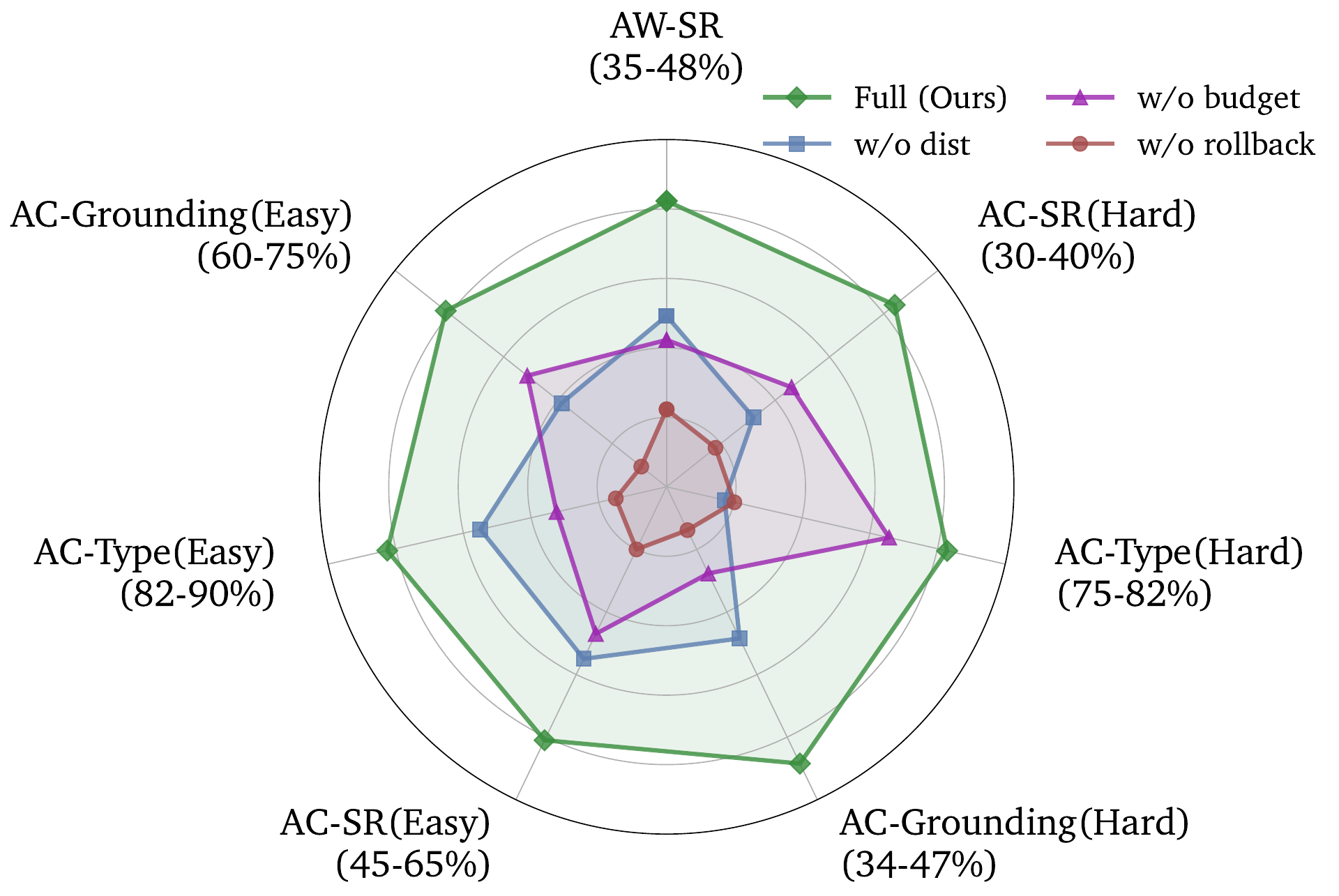}
    \caption{Effectiveness of critical components of \texttt{MobileGen}. AndroidWorld is short for ``AW'', and AndroidControl-Curated is short for ``AC''.}
    \label{fig:ablation}
\end{figure}

\noindent\textbf{GUIOdyssey.} To assess the OOD transferability of knowledge from our synthesized data to more complex scenarios, we conduct experiments on GUI-Odyssey~\cite{lu2025guiodyssey}, which involves diverse screen sizes and device types entirely unseen during the training stage. Evaluation covers High-Level (HL) natural requests for real-world needs and Low-Level (LL) fine-grained tasks for precise, unambiguous guidance. As shown in Table~\ref{tab:gui-odyssey-annotated}, \texttt{MobileGen} achieves the most significant performance gains across all backbones. Regarding overall performance, Qwen3-VL-8B-Instruct fine-tuned with \texttt{MobileGen} achieves an average 1.46$\times$ improvement over its zero-shot performance. Remarkably, our model's overall performance exceeds that of all specialized GUI Agents (i.e., UI-TARS-1.5-7B~\cite{qin2025ui} and GUI-Owl-7B~\cite{ye2025mobile}. These results demonstrate that difficulty-aware synthetic data effectively bolsters model generalization, ensuring robust cross-app interaction and stability across diverse OOD environments.

\begin{table*}[t]
\centering
\renewcommand{\arraystretch}{1.0} 
\small
\caption{Results on GUI-Odyssey. ``HL'' and ``LL'' denote high-level and low-level task instructions, respectively. The metric reported is the action match score (AMS). Results in \textbf{\hlt{A}{co}\hlt{B}{lo}\hlt{C}{red}} highlights and \underline{underline} denote the best and second-best results of each trained backbone, respectively.}
\resizebox{0.98\textwidth}{!}{
\begin{tabular}{l ccccc ccccc cccc}
\toprule
\multirow{2}{*}{\textbf{Model}} &
\multicolumn{2}{c}{\textbf{Tool}} &
\multicolumn{2}{c}{\textbf{Information}} &
\multicolumn{2}{c}{\textbf{Shopping}} &
\multicolumn{2}{c}{\textbf{Media}} &
\multicolumn{2}{c}{\textbf{Social}} &
\multicolumn{2}{c}{\textbf{Multi-Apps}} &
\multicolumn{2}{c}{\textbf{Overall}} \\
\cmidrule(lr){2-3}\cmidrule(lr){4-5}\cmidrule(lr){6-7}\cmidrule(lr){8-9}\cmidrule(lr){10-11}\cmidrule(lr){12-13}\cmidrule(lr){14-15}
& \textbf{HL} & \textbf{LL} & \textbf{HL} & \textbf{LL} & \textbf{HL} & \textbf{LL} & \textbf{HL} & \textbf{LL} & \textbf{HL} & \textbf{LL} & \textbf{HL} & \textbf{LL} & \textbf{HL} & \textbf{LL} \\
\midrule
GPT-5 & 38.6 & 75.5 & 27.7 & 65.4 & 23.9 & 61.0 & 26.9 & 71.4 & 36.6 & 74.8 & 24.4 & 65.5 & 29.2 & 68.4 \\
GUI-Owl-7B & 71.3 & 73.0 & 58.1 & 62.0 & 52.6 & 68.4 & 59.7 & 67.9 & 66.5 & 71.0 & 54.1 & 64.0 & \underline{59.7} & 66.9 \\
UI-TARS-1.5-7B & 52.7 & 66.0 & 39.8 & 54.7 & 31.6 & 49.1 & 47.5 & 61.6 & 48.8 & 64.8 & 39.2 & 54.1 & 42.7 & 57.7 \\
\midrule
\rowcolor{1} \multicolumn{15}{l}{\textbf{\textit{Zero-shot}}} \\

Qwen3-VL-4B-Inst. & 44.5 & 67.8 & 37.1 & 57.8 & 33.3 & 55.7 & 36.3 & 59.3 & 42.9 & 67.1 & 33.4 & 57.6 & 37.5 & 60.5 \\
Qwen3-VL-8B-Inst. & 44.7 & 70.8 & 35.3 & 59.6 & 31.2 & 58.5 & 36.0 & 63.0 & 41.4 & 69.1 & 34.4 & 60.5 & 37.0 & 63.2 \\
InternVL3-14B & 23.3 & 44.4 & 18.6 & 41.4 & 18.1 & 38.2 & 15.1 & 34.6 & 22.5 & 43.6 & 18.3 & 37.8 & 19.4 & 40.1 \\

\midrule
\rowcolor{1} \multicolumn{15}{l}{\textbf{\textit{OS-Genesis}}} \\
Qwen3-VL-4B-Inst. & 55.1 & 76.4 & 46.4 & 64.0 & \underline{47.3} & 66.2 & 47.8 & 68.3 & \underline{58.2} & 76.1 & 50.2 & 69.9 & 49.0 & 70.0 \\
Qwen3-VL-8B-Inst. & 60.6 & 78.8 & 51.0 & 67.0 & 49.4 & 69.1 & 52.1 & 74.1 & 61.1 & 79.6 & 56.0 & 74.4 & 53.6 & 73.3 \\
InternVL3-14B & \hlt{C}{\textbf{62.8}} & 75.1 & \hlt{C}{\textbf{52.8}} & \underline{66.1} & 46.7 & \underline{68.5} & 48.5 & 70.3 & 59.4 & 78.0 & 53.8 & 72.8 & 52.1 & 71.0 \\

\midrule
\rowcolor{1} \multicolumn{15}{l}{\textbf{\textit{GUI-ReWalk}}} \\
Qwen3-VL-4B-Inst. & \underline{57.2} & \hlt{A}{\textbf{78.1}} & 48.2 & \hlt{A}{\textbf{66.2}} & 47.0 & \hlt{A}{\textbf{71.9}} & \underline{51.4} & 74.4 & 56.2 & \underline{79.3} & \underline{52.2} & \underline{72.2} & 51.8 & \underline{72.1} \\
Qwen3-VL-8B-Inst. & \hlt{B}{\textbf{63.9}} & \hlt{B}{\textbf{80.5}} & \underline{53.9} & \hlt{B}{\textbf{69.8}} & \underline{51.1} & \underline{71.0} & \underline{56.9} & \hlt{B}{\textbf{78.6}} & \underline{63.0} & \hlt{B}{\textbf{82.5}} & \underline{58.0} & \underline{76.9} & \underline{56.8} & \underline{75.9} \\
InternVL3-14B & 59.9 & 77.6 & 50.5 & 65.5 & 48.2 & 67.2 & \hlt{C}{\textbf{53.6}} & \underline{72.3} & \underline{60.5} & \hlt{C}{\textbf{83.4}} & \underline{55.2} & \underline{74.0} & \underline{53.4} & \underline{72.4} \\

\midrule
\rowcolor{2} \multicolumn{15}{l}{\textbf{\textit{Ours}}} \\
Qwen3-VL-4B-Inst. & \hlt{A}{\textbf{58.9}} & \underline{77.7} & \hlt{A}{\textbf{49.5}} & 66.1 & \hlt{A}{\textbf{49.1}} & \underline{70.8} & \hlt{A}{\textbf{52.1}} & \hlt{A}{\textbf{76.3}} & \hlt{A}{\textbf{60.2}} & \hlt{A}{\textbf{81.1}} & \hlt{A}{\textbf{54.2}} & \hlt{A}{\textbf{74.2}} & \hlt{A}{\textbf{54.8}} & \hlt{A}{\textbf{75.1}} \\
Qwen3-VL-8B-Inst. & \underline{63.5} & \underline{79.9} & \hlt{B}{\textbf{53.5}} & \underline{69.1} & \hlt{B}{\textbf{53.1}} & \hlt{B}{\textbf{72.7}} & \hlt{B}{\textbf{58.0}} & \underline{76.9} & \hlt{B}{\textbf{64.2}} & \underline{80.3} & \hlt{B}{\textbf{60.0}} & \hlt{B}{\textbf{78.9}} & \hlt{B}{\textbf{59.8}} & \hlt{B}{\textbf{81.3}} \\
InternVL3-14B & \underline{61.1} & \hlt{C}{\textbf{78.9}} & \underline{52.1} & \hlt{C}{\textbf{67.8}} & \hlt{C}{\textbf{50.2}} & \hlt{C}{\textbf{70.4}} & \underline{52.5} & \hlt{C}{\textbf{74.3}} & \hlt{C}{\textbf{62.1}} & \underline{82.9} & \hlt{C}{\textbf{57.2}} & \hlt{C}{\textbf{76.0}} & \hlt{C}{\textbf{56.4}} & \hlt{C}{\textbf{75.4}} \\

\bottomrule
\end{tabular}
}

\label{tab:gui-odyssey-annotated}
\end{table*}

\subsection{Deep Dive}
\noindent\textbf{Effectiveness of Critical Components.} \label{exp:ablation}
We systematically investigate 
the core components of \texttt{MobileGen} from two primary perspectives: \textbf{(1) Customized Difficulty Distribution}, by replacing our profiling-driven distribution with random difficulty selection (\texttt{MobileGen} w/o \textit{dist}); and \textbf{(2) Difficulty-Aware Generation Control}, by independently disabling the \textit{step budget} and \textit{rollback} mechanisms (\texttt{MobileGen} w/o \textit{budget}/\textit{rollback}) to assess their necessity in difficulty regulation.

As illustrated in Figure~\ref{fig:ablation}, removing any core component results in a significant performance degradation across both benchmarks. Notably, \texttt{MobileGen} w/o \textit{rollback} exhibits the most substantial decline, with a performance gap of 4.4\%-12.2\% compared to the full version of \texttt{MobileGen}. This validates that the complete \texttt{MobileGen} pipeline--from agent capability profiling to difficulty-aware data generation--is indispensable for effectively enhancing the autonomous interaction capabilities of VLMs in complex GUI environments.


\begin{figure}
    \centering
    \includegraphics[width=1\linewidth]{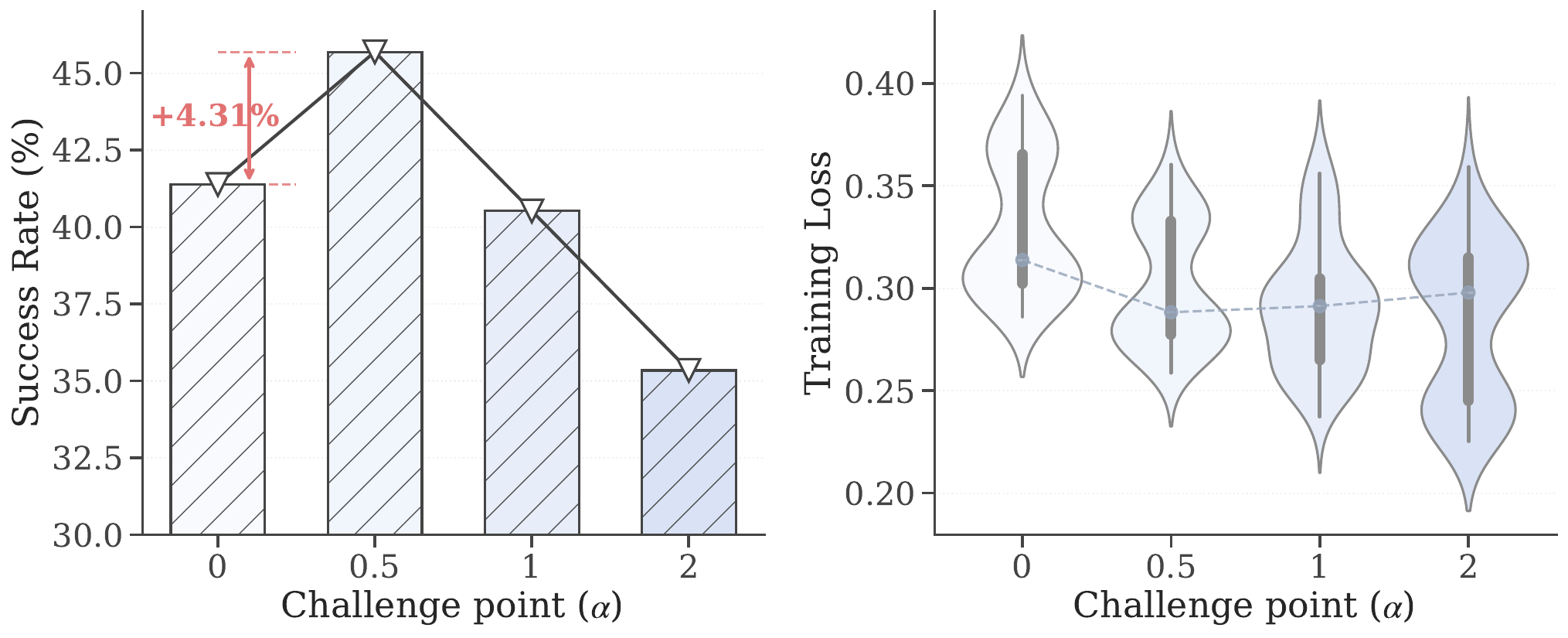}
    \caption{Evaluation of training under different challenge point setting. Success rate and training loss distribution are reported.}
    \label{fig:point}
\end{figure}

\textbf{Impact of Challenge Point Setting.}
As the agent capability acquisition is intrinsically linked to the setting of the challenge point, we systematically evaluate the influence of the challenge point ($\alpha$) by training Qwen3-VL-8B-Instruct~\cite{bai2025qwen3VL} across a spectrum from $\alpha$ = 0 (trivial) to $\alpha$ = 2 (hard). As illustrated in Figure \ref{fig:point}, the $\alpha$ = 0.5 setting yields the optimal learning efficiency, outperforming other settings by a margin of 4.3\% and achieving the minimum training loss. This trend underscores that either trivial or excessively difficult tasks result in suboptimal learning. Such evidence emphasizes the effectiveness of our difficulty-aware framework, which enhances learning by aligning data generation around the agent’s current capability boundary.

\noindent\textbf{Data Analysis.} 
To evaluate the quality of trajectories synthesized by \texttt{MobileGen}, we also compare its generated trajectories against those generated by baselines through quantitative and visual analysis in  Appendix \ref{apdx:sup_exp}. Our analysis shows that \texttt{MobileGen} generates diverse trajectories spanning a range of difficulty levels and consistently outperforms baselines in terms of overall data quality.






\section{Conclusion}
\label{conclusion}

In this paper, we present \texttt{MobileGen}, an adaptive data generation framework designed to address the shortage of high-quality training data for mobile GUI agents. By shifting from stochastic collection toward a difficulty-aware paradigm, \texttt{MobileGen} maximizes learning effectiveness by precisely aligning task difficulty with an agent’s current capability frontier. 
Our framework decouples trajectory difficulty into structural and semantic dimensions, and iteratively (1) profiles the agent’s capability, (2) sets the challenge point for training, and (3) synthesizes high-quality data through multi-agent collaboration.
Empirical evaluations across three benchmarks demonstrate that  
\texttt{MobileGen} consistently yields superior success rates and robust generalization. Notably, by leveraging only 500 high-quality synthetic trajectories, \texttt{MobileGen} significantly bridges the gap between general-purpose VLMs and SOTA specialized GUI agents, maintaining a robust advantage in complex OOD scenarios. 
These results underscore the critical importance of difficulty-aware data synthesis in advancing the capabilities of the next generation of autonomous digital assistants.

\section*{Impact Statement}


This paper presents work whose goal is to advance the field of Machine
Learning. There are many potential societal consequences of our work, none
which we feel must be specifically highlighted here.



\section*{Author Contributions}
\begin{itemize}
    \item Linjia Kang initiated and led the project. She is responsible for the entire research process from motivation formulation and methodology design to experiment design. She led the implementation of the \texttt{MobileGen} and completed the majority of the manuscript writing and revisions. 
    \item Zhimin Wang made significant contributions to the methodology design and optimization strategies of the project. He is also responsible for a substantial portion of the manuscript writing, revision, and figure refinement. 
    \item Yongkang Zhang reproduced the baseline methods and contributed to the partial implementation of \texttt{MobileGen}. He also assisted with system performance optimization and conducted large-scale experiments. 
    \item Duo Wu provided extensive supervision throughout the project, including guidance on problem formulation, idea refinement, experimental design, and manuscript writing, and also contributed to the paper revision. 
    \item Jinghe Wang, Ming Ma, and Haopeng Yan provided technical suggestions and support. 
    \item Zhi Wang provided overall supervision, institutional resources, and financial support for the project.
\end{itemize}

\bibliography{bibfile/gui}
\bibliographystyle{icml2026}

\newpage
\appendix
\onecolumn


\section{Appendix}
\subsection{Data Analysis}
\label{apdx:sup_exp}

\subsubsection{Comparison of Data Quality.}
To evaluate the quality of the synthesized data, we employ advanced VLM judge to assess two distinct dimensions: step-level quality and trajectory-level quality. 
For a fair comparison, we evaluate a fixed sample of 0.5k trajectories per method, all generated using the same Qwen3-VL-8B-Instruct~\cite{bai2025qwen3VL} backbone. 
As illustrated in Table~\ref{tab:data_quality}, \texttt{MobileGen} consistently achieves superior scores across both granularities. This advantage remains robust across varying difficulty profiles derived from three different VLMs. Details of the quality assessment are provided in Appendix~\ref{apdx:data_quality}.



\begin{table}[ht]
\centering
\renewcommand{\arraystretch}{1.0} 
\begin{small} 
\caption{Comparison of trajectory quality across different methods. The model names in the brackets denote the student agents that provide difficulty profiles. We report the average quality score of \texttt{MobileGen} across three models.}
\label{tab:data_quality}
\begin{tabularx}{0.95\textwidth}{lcccc} 
\toprule
\textbf{Datasets} & \textbf{Average Steps} & \textbf{Average App number} & \textbf{Step-level Quality} & \textbf{Trajectory-level Quality} \\
\midrule
OS-genesis        & 20.37 & 1.37 & 6.73 & 5.84 \\
GUI-ReWalk       & 21.35 & 1.83 & 7.39 & 7.10 \\
\midrule
Ours (Qwen3-VL-4B-Inst.)   & 17.19 & 1.54 & \textbf{8.37}   & 8.19   \\
Ours (Qwen3-VL-8B-Inst.)   & 21.51 & 2.35 & 8.13 & 8.30 \\
Ours (InternVL3-14B) & 18.50 & 1.86 & 8.14 & \textbf{8.44} \\
\bottomrule
\end{tabularx}
\end{small}
\vspace{2mm}
\end{table}

\subsubsection{Visualization Analysis}
\begin{figure}[h]
    \centering
    \includegraphics[width=1.0\linewidth]{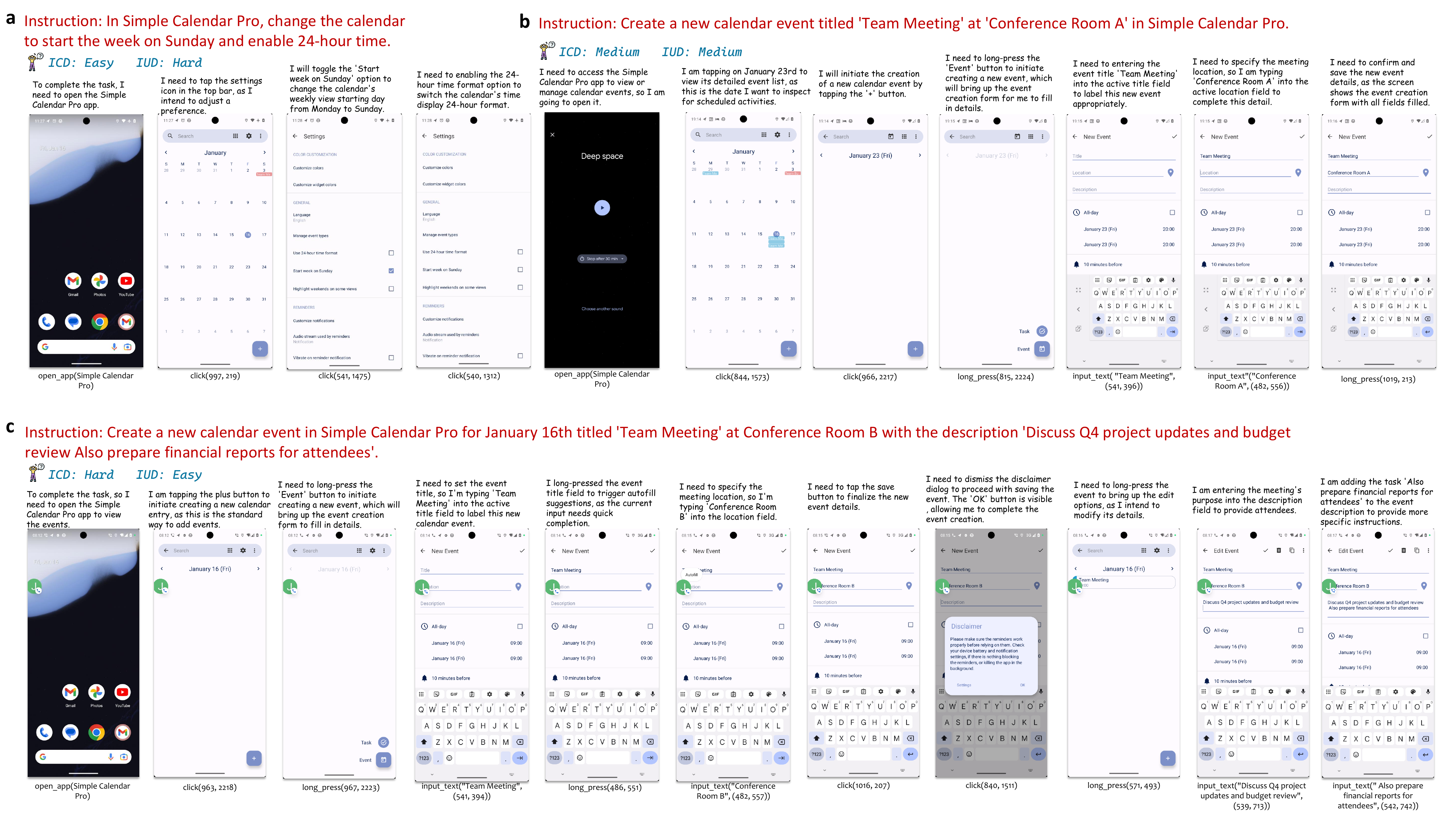}
    \caption{Case study of semantic difficulty controllability. By visualizing trajectories generated in Simple Calendar Pro, we demonstrate the effectiveness of adjusting ICD and IUD parameters to control semantic complexity: (a) illustrates a setting modification task with low operational complexity and an ambiguous instruction; (b) presents a standard event creation task with medium difficulty in both linguistic logic and interaction; and (c) showcases a more complex event creation task that demands intricate operational requirements but straightforward instruction understanding.}
    \label{fig:semantic_case}
\end{figure}

To empirically evaluate the semantic controllability of our framework, we conducted a detailed case study using Simple Calendar Pro. By aligning the application context across three trajectories, we eliminated environmental interference, allowing for a more precise isolation and comparison of how different ICD and IUD settings influence trajectory complexity.
As Figure~\ref{fig:semantic_case} shows, 
trajectory (a) simulates a scenario with a ambiguous instruction, requiring the model to explore the interface based on intent understanding. Although the operational logic is simple—only requiring the model to click common toggle buttons—the high IUD forces the agent to implicitly infer user intent and navigate through hierarchical menus, proving our framework's ability to control path depth even for simple goals.
Trajectory (b) illustrates a standard event creation process. Both its linguistic logic and interaction steps are at a medium level, reflecting the typical difficulty found in common user-agent interactions.
Trajectory (c) showcases the framework's capacity for high-density information filling. The task involves multiple detailed fields, including title, location, and long-form descriptions. Notably, the visualization reveals that the trajectories may follow suboptimal paths, involving some redundant interface transitions. Yet, combined with our experimental results in Section~\ref{exp:main_results}, we find that these successful yet suboptimal trajectories can still effectively train GUI agents, as they may provide the behavioral diversity necessary for learning.
This case study demonstrates the capability of \texttt{MobileGen} to steer the synthesis of data with varying semantic difficulties. Furthermore, even with the presence of suboptimal behaviors, the resulting synthetic dataset significantly enhances the decision-making and grounding capabilities of VLMs in GUI environments.

\clearpage

\subsection{Details of Benchmarks}
\label{apdx:benchmark}

\noindent \textbf{AndroidWorld.} AndroidWorld~\cite{rawles2024androidworld} is a dynamic, fully functional benchmarking environment designed for evaluating autonomous agents on the Android platform. It features 116 programmatic tasks spanning 20 real-world applications, including productivity, communication, and system utilities. Unlike static datasets that rely on matching human demonstrations, AndroidWorld dynamically instantiates tasks using randomized parameters and natural language instructions, allowing for millions of unique task goals to ensure evaluation robustness. The environment leverages the Android Emulator and the Python library AndroidEnv to facilitate asynchronous interaction. Success is verified through a non-invasive reward mechanism that inspects the underlying system state—such as SQLite databases, file systems, and system settings—ensuring durable and accurate evaluation without requiring modifications to the application source code. \klj{The benchmark evaluates agents based on the Success Rate (labeled as SR), which is defined as the percentage of tasks where the agent successfully achieves the stated goal.}

\noindent \textbf{AndroidControl-Curated.} AndroidControl-Curated~\cite{leung2025androidcontrol} is a purified version of the original AndroidControl~\cite{li2024effects} dataset, designed to rectify systemic flaws that lead to the underestimation of GUI agent capabilities. The benchmark was developed through a rigorous semi-automated pipeline that addresses two primary areas: evaluation-level bias and data-level factual errors. The dataset was created by filtering the original AndroidControl (which consists of approximately 90k samples) to identify and correct high-risk entries. \klj{The benchmark utilizes three primary metrics to assess agent performance: 
(1) Success Rate (labeled as SR) is the ultimate metric for evaluating an agent's ability to complete a task; a task is considered successful only if the agent executes all steps correctly and reaches the intended final screen state. 
(2) Grounding Accuracy (labeled as Grounding) measures the agent's ability to localize the target for an operation at each step, calculated using the intent-aligned evaluation function:
\begin{equation}
    E_{bbox}(p_{pred}, p_{gt}) = \mathbb{I}(p_{pred} \in B_{gt}) = \mathbb{I}(p_{pred} \in \mathcal{F}(p_{gt}))
\end{equation}
where $p_{pred}$ is the predicted point, $B_{gt}$ is the ground-truth interaction area, and $\mathcal{F}$ is a mapping function to the minimal UI element's bounding box. 
(3) Action-type Accuracy (labeled as Type) evaluates the correctness of the predicted action type (e.g., \textit{Click}, \textit{Type}, \textit{Scroll}) for each step in a trajectory.}

\noindent \textbf{GUIOdyssey.} GUIOdyssey~\cite{lu2025guiodyssey} is the first comprehensive dataset specifically designed for cross-app GUI navigation on mobile devices, addressing the limitations of prior benchmarks that focus primarily on single-app tasks. The dataset comprises 8,334 episodes with an average length of 15.3 steps, the longest among existing mobile GUI datasets, covering 212 unique applications and 1,357 application combinations across six device types including tablets and foldables. Tasks are categorized into six types: General Tool, Information Management, Web Shopping, Media Entertainment, Social Sharing, and Multi-Apps. Each step features dual-level instructions—high-level natural language requests and fine-grained low-level instructions—alongside rich semantic annotations including screen descriptions, contextual history summaries, and decision rationales to facilitate complex reasoning. \klj{The primary metric is the Action Matching Score (AMS), which evaluates step-level correctness. An action is correct if its type matches the ground truth and, for \texttt{Click} or \texttt{Long Press}, the coordinates fall within 14\% of the screen distance from the gold gesture or within the region segmented by SAM2~\cite{ravi2025sam2}. \texttt{Scroll} actions must match the gold direction, and \textit{Type} actions must exceed an Average Normalized Levenshtein Similarity~(ANLS) threshold of 0.5.}


\clearpage

\subsection{Experimental Details}\label{apdx:exp_details}
\subsubsection{Action Space}
The dataset generated by \texttt{MobileGen} exclusively contains actions that fall under the categories specified in Table~\ref{tab:action_space_mobile}.
In the case of AndroidWorld~\cite{rawles2024androidworld}, three extra actions—\texttt{terminate}, \texttt{answer}, and \texttt{status}—are added to fulfill the evaluation requirements.

\begin{table}[h!]
\centering
\caption{Action space for \texttt{MobileGen}.}
\label{tab:action_space_mobile}
\begin{tabular}{ll}
\hline
\textbf{Action} & \textbf{Description} \\
\hline
click & Performs a tap on the selected UI element. \\
long\_press & Holds a tap on the specified element for an extended duration. \\
scroll & Moves the visible portion of the screen or UI element in the indicated direction. \\
input\_text & Enters the given text into the chosen text field at the cursor position. \\
navigate\_home & Returns the device to the main home screen. \\
navigate\_back & Goes back to the previous screen or page within the app. \\
open\_app & Starts the designated application. \\
wait & Pauses the agent’s action until the screen is ready. \\
keyboard\_enter & Sends the Enter key input. \\
\hline
terminate & Marks the current task as completed by the agent. \\
answer & Outputs a response, typically used for retrieving requested information. \\
status & Indicates whether the task is ongoing, finished, or cannot be completed. \\
\hline
\end{tabular}
\end{table}

\subsubsection{Data Quality Assessment}\label{apdx:data_quality}
\noindent\textbf{Step-Level Quality.} Our evaluation focuses on the microscopic integrity of each individual interaction. We assess this through two primary dimensions: \textit{Grounding} and \textit{Goal Alignment}. Grounding measures the internal consistency between the agent's reasoning and the physical action executed, ensuring that the agent is not ``hallucinating" its logic but is instead accurately interacting with the intended UI elements. Goal Alignment, on the other hand, evaluates the external relevance of the step, verifying whether the action provides a plausible contribution toward the high-level task instruction based on the visual evidence before and after the move. The detailed scoring rubric and evaluation criteria are illustrated in Figure~\ref{fig:step_quality_prompt}.

\noindent\textbf{Trajectory-Level Quality.} We shift to a macroscopic view to evaluate the overall effectiveness of the interaction sequence. This level likewise employs two dimensions: \textit{Goal Achievement} and \textit{Step Efficiency}. The Goal dimension assesses the final outcome—specifically whether the entire sequence results in successful task completion or meaningful progress toward the objective. Parallelly, the Efficiency dimension monitors the quality of the path taken, penalizing redundant actions, wasted steps, or repetitive loops, aiming to faithfully reflect the true utility and directness of the trajectory. The corresponding evaluation prompt and trajectory-level dimensions are presented in Figure~\ref{fig:traj_quality_prompt}.

\begin{figure*}[h]
\centering
\begin{tcolorbox}[
    colback=myback,        
    colframe=myframe,  
    listing only, 
    title=\textbf{\textit{Evaluation Prompt for Step-Level Quality}}
]
You are an expert judge for SINGLE-STEP Android GUI agent data.\par
\vspace{0.5em}
You will be given: \\
- High-level Instruction (the overall task for this trajectory) \\
- Action JSON (the action executed) \\
- Agent Reasoning (why the agent claims it chose this action; may be empty) \\
- Target element info (if any) \\
- BEFORE screenshot and AFTER screenshot (raw screenshots; no UI boxes/SoM marks are required)\par
\vspace{0.5em}
You MUST evaluate the step using 2 internal dimensions: \par
1) \textbf{Grounding (action $\leftrightarrow$ reasoning alignment):} \\
- Does the reasoning correctly describe and justify the given action? \\
- Penalize contradictions, wrong targets, or hallucinated claims that do not match the action. \par
2) \textbf{Goal Alignment (action $\leftrightarrow$ high-level instruction alignment):} \\
- Does the action plausibly advance the High-level Instruction, given the BEFORE/AFTER evidence? \\
- Not every step must directly complete the task, but it should be coherent progress or necessary navigation. \\
- Penalize actions that are unrelated, clearly wrong, or contradict the high-level task.\par
\vspace{0.5em}
You must output ONE unified score (integer 1-10) that reflects BOTH dimensions. \par
IMPORTANT: the unified score should be close to the WEAKER of the two dimensions. If either grounding or goal alignment is low, the overall score must be low.\par
\vspace{0.5em}
Scoring tiers (be conservative; when uncertain, choose the LOWER score): \\
- 10: grounding AND goal alignment are both excellent; no obvious mismatch; strong training signal. \\
- 8-9: both mostly correct; minor ambiguity or minor inefficiency; still reliable. \\
- 6-7: partial alignment; noticeable ambiguity/mismatch in at least one dimension; still usable. \\
- 4-5: weak signal; frequent mismatch, confusion, or limited evidence of goal progress. \\
- 1-3: clearly wrong/contradictory; very weak or misleading training signal.\par
\vspace{0.5em}
Output format (strict): \\
- Output ONLY one JSON object (no extra text). \\
- The JSON MUST have exactly these keys: \\
\hspace*{1em} \{ \\
\hspace*{2em} "score": integer 1-10, \\
\hspace*{2em} "reason": "a short sentence explaining the main issue(s)" \\
\hspace*{1em} \}
\end{tcolorbox}
\caption{Evaluation prompt for step-level quality.}
\label{fig:step_quality_prompt}
\end{figure*}

\begin{figure*}[h]
\centering
\begin{tcolorbox}[
    colback=myback,        
    colframe=myframe,  
    listing only, 
    title=\textbf{\textit{Evaluation Prompt for Trajectory-Level Quality}}
]

You are an expert judge for Android GUI agent trajectories. Your job is to output ONE final reward score from 1 to 10.\par
\vspace{0.5em}
You MUST evaluate the trajectory using 2 internal dimensions: \\
1) \textbf{Goal Achievement:} task completion / progress toward the high-level instruction \\
2) \textbf{Step Efficiency:} redundancy / loops / wasted steps (DO NOT penalize repetition if the instruction explicitly asks for it)\par
\vspace{0.5em}
\textbf{IMPORTANT:} \\
- Do NOT score step-level grounding here (e.g., whether a specific action is supported by UI evidence). Assume step-level grounding is evaluated separately.\par
\vspace{0.5em}
You MUST follow the output format exactly: \\
- Output ONLY TWO LINES, no extra text, no markdown, no code blocks. \\
- Line 1 MUST start with: Reason: \\
- Line 2 MUST start with: Score: \\
- Score MUST be an integer in [1, 10].\par
\vspace{0.5em}
Scoring guidance (be conservative; when uncertain, choose the LOWER score): \\
- Score 10: achieves the goal; efficient; no obvious loops. \\
- Score 8-9: strong coherent progress; minor inefficiency/redundancy. \\
- Score 6-7: partial progress; noticeable inefficiency or mild confusion; still useful. \\
- Score 3-5: little progress and/or heavy redundancy; low-value trajectory. \\
- Score 1-2: deadlock/loop, almost no useful signal. \par
\vspace{0.5em}
High-level Instruction: \$TASK\_INSTRUCTION\$ \\
Trajectory Stats (computed): \$STATS\$ \\
Latest Steps: \$LATEST\_STEPS\$ \\
The last \$NUM\_SCREENSHOTS\$ screenshots are provided.\par
\vspace{0.5em}
\textbf{FORMAT REMINDER:}\par
\vspace{0.5em}
Output ONLY TWO LINES: \\
Reason: $<$short$>$ \\
Score: $<$integer 1-10$>$

\end{tcolorbox}
\caption{Evaluation prompt for trajectory-level quality.}
\label{fig:traj_quality_prompt}
\end{figure*}

\clearpage

\subsubsection{Hyperparameter Settings}
To ensure the reproducibility of our experiments, we provide a comprehensive overview of the hyperparameter settings used in this study. The parameters are categorized into two parts:

\noindent\textbf{\texttt{MobileGen}.} Table 1 details the default settings for our \texttt{MobileGen} framework, including the evolutionary factors ($\alpha$, $\eta$) and sampling distribution widths ($\sigma$) mentioned in Section~\ref{sec:distribution_generation}, and temprature for VLMs.

\noindent\textbf{Model Training.} Table 2 summarizes the specific hyperparameters for the SFT process, covering LoRA adapter settings, optimization schedules, and hardware-related acceleration configurations.

\begin{table}[h]
\centering
\caption{Default hyperparameter settings for \texttt{MobileGen}.}
\label{tab:hyperparameter_ours}
\small
\setlength{\tabcolsep}{4pt}   
\renewcommand{\arraystretch}{1.25}
\begin{tabularx}{\textwidth}{|c|X|c|}
\hline
\textbf{Parameter} & \textbf{Description} & \textbf{Value} \\
\hline
$\alpha$ &
Challenge increment factor controlling how far the target difficulty advances beyond the agent’s estimated capability frontier &
$0.5$ \\
\hline
$\eta_d$ &
Scaling factor for advancing sequential execution capability relative to $\alpha$ &
$6.0$ \\
$\eta_b$ &
Scaling factor for advancing cross-application interaction capability relative to $\alpha$ &
$1.0$ \\
$\eta_{\text{int}}$ &
Scaling factor for advancing interaction control capability relative to $\alpha$ &
$0.8$ \\
$\eta_{\text{ins}}$ &
Scaling factor for advancing instruction understanding capability relative to $\alpha$ &
$0.8$ \\
\hline
$\sigma_d$ &
Gaussian kernel width controlling the concentration of the DoT sampling distribution &
$3$ \\
$\sigma_b$ &
Gaussian kernel width controlling the concentration of the BoT sampling distribution &
$0.5$ \\
$\sigma_a$ &
Gaussian kernel width controlling the application selection distribution &
$1.0$ \\
\hline
$k$ &
Number of samples used in Pass@K for agent capability profiling &
$3$ \\
\hline
$\beta_{\text{gen}}$ &
Softmax temperature controlling model stochasticity during data generation &
$0.0$ \\
$\beta_{\text{prof}}$ &
Softmax temperature controlling model stochasticity during capability profiling &
$0.2$ \\
\hline
\end{tabularx}
\end{table}


\begin{table}[ht]
\centering
\caption{Hyperparameter settings for the model training.}
\label{tab:hyperparameters_training}
\small
\setlength{\tabcolsep}{4pt}   
\renewcommand{\arraystretch}{1.25}
\begin{tabularx}{\textwidth}{|l|X|c|}
\hline
\textbf{Parameter} & \textbf{Description} & \textbf{Value} \\
\hline
\texttt{lora\_rank} & Rank of the LoRA & 32 \\
\texttt{lora\_alpha} & Scaling factor for LoRA & 64 \\
\texttt{lora\_dropout} & Dropout probability for LoRA & 0.05 \\
\hline
\texttt{optimizer} & Optimizer type & paged\_adamw\_8bit \\
\texttt{learning\_rate} & Learning rate for optimization & 1e-4 \\
\texttt{lr\_scheduler\_type} & Learning rate decay schedule & cosine \\
\texttt{warmup\_ratio} & Fraction of total steps for linear warmup & 0.03 \\
\texttt{weight\_decay} & Weight decay coefficient for regularization & 0.0 \\
\texttt{max\_grad\_norm} & Maximum gradient norm for clipping & 1.0 \\
\texttt{gradient\_accumulation\_steps} & Number of steps to accumulate before backpropagation & 16 \\
\hline
\texttt{num\_epochs} & Total number of training epochs & 10 \\
\texttt{batch\_size} & Batch size per processing unit & 1 \\
\hline
\texttt{cutoff\_len} & Maximum sequence length for input tokens & 2048 \\
\texttt{image\_resolution} & Resolution for input image processing & 448 \\
\texttt{precision} & Floating-point precision & bf16 \\
\hline
\end{tabularx}
\end{table}

\clearpage

\subsection{Difficulty Parameters Sampling}\label{apdx:sampling}
For a trajectory $\tau$, its difficulty is characterized by four parameters: DoT denoted as $d_{\tau}$, BoT denoted as $b_{\tau}$, ICD denoted as $c_{\tau}$, and IUD denoted as $u_{\tau}$, which are defined in Section~\ref{sec:preliminaries}. To generate trajectories with appropriately controlled difficulty, we design a hybrid sampling procedure to sample trajectory parameters. 
Specifically, $d_{\tau}$, $c_{\tau}$, and $u_{\tau}$ are directly sampled from the corresponding probability distributions defined in Section~\ref{sec:distribution_generation}. 
In contrast, $b_{\tau}$ is sampled in a conditional manner: given $d_{\tau}$, its feasible range is explicitly truncated to satisfy $b_{\tau} \le d_{\tau}$, thereby enforcing the inherent inequality constraint at the sampling stage.
Furthermore, we adpot the mean relative failure frequency $\mathcal{V}^{\star} = \frac{1}{N_a} \sum_{i=1}^{N_a} \tilde{\mathcal{V}}_i$ to construct the application selection sampling distribution $p_{\text{app}}(z)$ using a gaussian kernel centered at $\mathcal{V}^{\star}$ similar to Equation~\ref{eq:gaussian}.
Under this formulation, applications whose $\tilde{\mathcal{V}}_i$ are closer to $\mathcal{V}^{\star}$—and thus correspond to tasks of moderate difficulty for the current agent—are assigned higher sampling probabilities. 
Each trajectory is generated by sampling its parameters from our predefined distributions. As the distributions are specified in advance, our pipeline naturally supports large-scale parallel trajectory sampling, enabling efficient construction of the training task $\mathscr{T}$ while maintaining statistical consistency with the intended difficulty distributions.

\subsection{Prior Dataset}\label{apdx:prior}
\subsubsection{Dataset Construction}
\begin{figure}[h]
    \centering
    \includegraphics[width=1.0\linewidth]{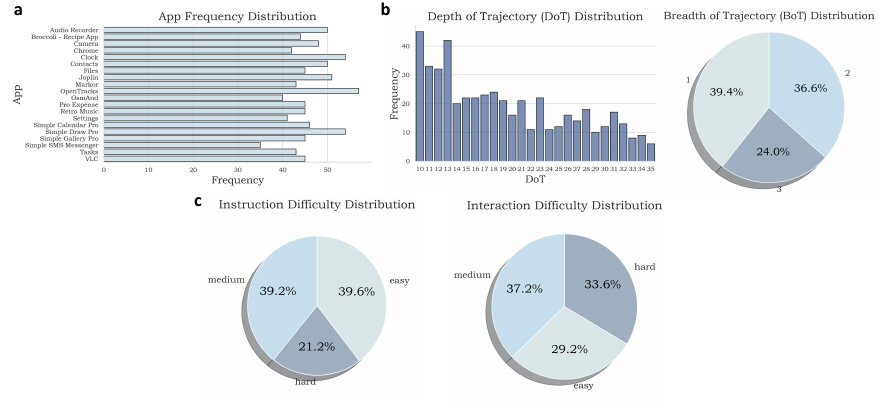}
    \caption{Statistics of the prior trajectory dataset $\mathscr{T}_p$.
    (a) Application frequency distribution, showing balanced coverage across diverse applications.
    (b) Structural difficulty statistics of trajectories, including the distribution of DoT and BoT.
    (c) Semantic difficulty distributions for IUD and ICD.
    Overall, the dataset exhibits a well-balanced distribution across applications and both structural and semantic difficulty dimensions.}
    \label{fig:prior_statistics}
\end{figure}

This section specifies the detailed parameter configurations and sampling methodologies used to construct the prior dataset $\mathscr{T}_p$ via our proposed \texttt{MCG}. To ensure a balanced and diverse distribution of trajectory difficulty, we configure the trajectory difficulty parameters (DoT, BoT, ICD, and IUD) as follows.
Specifically, each difficulty parameter is stratified into three levels: \textit{easy}, \textit{medium}, and \textit{hard}. For the structural difficulty parameters, the configurations are defined as: 
(1) \textit{easy} corresponds to DoT $\in$ [10, 15] and BoT = 1; 
(2) \textit{medium} corresponds to DoT $\in$ [16, 25] and BoT = 2; 
(3) \textit{hard} corresponds to DoT $\in$ [26, 35] and BoT = 3. 
Regarding ICD and IUD, their values inherently span the qualitative categorical levels of easy, medium, and hard.
The dataset generation follows a hierarchical grid-enumeration strategy. We first align the levels of DoT and BoT to form a unified structural complexity index, which is then cross-combined with the ICD levels in a $3 \times 3$ grid to ensure comprehensive coverage of operational scenarios. Finally, during the reverse instruction synthesis phase, the IUD level for each trajectory is independently determined by sampling from $\{\textit{easy, medium, hard}\}$ following a uniform distribution. 
To guarantee the quality of the $\mathscr{T}_p$, we utilize VLM judge defined in Appendix~\ref{apdx:data_quality} and Figure~\ref{fig:traj_quality_prompt} to score the trajectories, retaining only those with a score exceeding 8. Ultimately, our generated $\mathscr{T}_p$ comprises 500 high-quality trajectories. As illustrated in Figure~\ref{fig:prior_statistics}, the dataset exhibits a highly balanced distribution of applications and difficulty.

\subsubsection{Action Matching Protocol}\label{apdx:action_matching}
To ensure a rigorous and reproducible evaluation, we define a unified matcher $\mathcal{M}(\mathbf{a}_{pred}, \mathbf{a}_{gt})$. An action is considered a match (i.e., $\mathcal{M}=1$) if it satisfies specific categorical criteria. Before formal matching, a canonicalization operator $\mathcal{C}(\cdot)$ and a semantic mapping $\mathcal{S}(\cdot)$ are applied to mitigate linguistic noise. Specifically, $\hat{a} = \mathcal{S}(a_{type})$ maps various aliases (e.g., \textit{tap}, \textit{touch}) to a closed set of canonical types $\mathcal{T} = \{ \text{click, scroll, input\_text, \dots} \}$, while $\mathcal{C}(s)$ transforms strings for \texttt{open\_app} and \texttt{input\_text} by removing non-alphanumeric characters and normalizing to lowercase.

\noindent \textbf{Spatial and Textual Criteria.} For coordinate-based interactions such as \texttt{click} and \texttt{long\_press}, we adopt a hierarchical spatial verification strategy. Let $p_{pred}$ and $p_{gt}$ be the interaction coordinates, and $B_{gt}$ be the target bounding box pixels. A match is granted if the predicted point resides within the ground-truth region or falls within a diagonal distance threshold:
\begin{equation}
    \mathcal{M} = \mathbb{I} \left( (p_{pred} \in B_{gt}) \lor \left( \frac{\| p_{pred} - p_{gt} \|_2}{\sqrt{W^2 + H^2}} \leq \phi \right) \right)
\end{equation}
where $\phi = 0.14$ is the tolerance ratio following GUIOdyssey~\cite{lu2025guiodyssey}. For \texttt{input\_text}, the validity is governed by ANLS, defined as $\text{ANLS}(T_{pred}, T_{gt}) = 1 - \text{Lev}(T_{pred}, T_{gt}) / \max(|T_{pred}|, |T_{gt}|)$. A successful match requires $\text{ANLS} \geq 0.5$ in addition to satisfying the aforementioned spatial criteria.

\noindent \textbf{Discrete and Navigation Rules.} For actions involving system-level triggers or discrete states, we enforce exact-match rules as summarized in Table~\ref{tab:action_match_rules}. The \texttt{scroll} action is matched strictly by the direction attribute $\text{Direction} \in \{up, down, left, right\}$, while \texttt{open\_app} requires an exact string match of canonicalized application names. Atomic actions such as \texttt{home}, \texttt{back}, and \texttt{wait} are matched solely by their canonical type identity $\hat{a}$.

\begin{table}[h]
\centering
\caption{Exact-match rules for discrete and navigation actions.}
\label{tab:action_match_rules}
\small
\renewcommand{\arraystretch}{1.2}
\begin{tabular}{ll}
\toprule
\textbf{Action Type} & \textbf{Matching Condition} \\
\midrule
\texttt{click}          & $p_{pred} \in B_{gt}$ or $\|p_{pred}-p_{gt}\|_2 / \sqrt{W^2+H^2} \leq \phi$ \\
\texttt{long\_press}    & $p_{pred} \in B_{gt}$ or $\|p_{pred}-p_{gt}\|_2 / \sqrt{W^2+H^2} \leq \phi$ \\
\texttt{scroll}          & $\text{Direction}_{pred} = \text{Direction}_{gt}$ \\
\texttt{input\_text}    & ANLS($T_{pred}, T_{gt}$) $\geq 0.5$ and the predicted point satisfies the above spatial condition \\
\texttt{open\_app}       & $\mathcal{C}(\text{App}_{pred}) = \mathcal{C}(\text{App}_{gt})$ \\
\texttt{navigate\_home}  & $\hat{a}_{pred} = \hat{a}_{gt}$ \\
\texttt{navigate\_back}  & $\hat{a}_{pred} = \hat{a}_{gt}$ \\
\texttt{wait}            & $\hat{a}_{pred} = \hat{a}_{gt}$ \\
\texttt{keyboard\_enter} & $\hat{a}_{pred} = \hat{a}_{gt}$ \\
\bottomrule
\end{tabular}
\end{table}

\clearpage

\subsection{Multi-Agent Controllable Generator}\label{apdx:mcg}
\subsubsection{Step Budget Management}
Algorithms~\ref{alg:budget_allocation} and~\ref{alg:weight_update} together constitute the step budget control mechanism.
Given a set of pending applications, Algorithm~\ref{alg:budget_allocation} allocates the total step budget by querying the supervisor model $\mathcal{M}_{\text{supervisor}}$.
When the allocation weights are not yet initialized, they are set equally for all applications. Otherwise, the existing weights are reused.
The supervisor then assigns exploration step budgets to individual applications conditioned on the current weight distribution.
Algorithm~\ref{alg:weight_update} updates the weights only when the cyclic interaction behavior of $\mathcal{M}_{\text{explorer}}$ within an application exceeds a predefined threshold $\gamma$.
In such cases, the weight of the corresponding application is penalized and re-normalized to suppress subsequent inefficient exploration of that application. If no cyclic behavior is detected, the weights remain unchanged.
This design achieves adaptive responses to exploration feedback while maintaining stability in budget allocation.
\subsubsection{Rollback}
The rollback mechanism dynamically monitors the global working memory $\mathcal{H}$ to detect anomalous interaction patterns within the current application.  
Upon detecting an anomaly, $\mathcal{M}_{\text{supervisor}}$ identifies the appropriate historical step for recovery and triggers a rollback.  
To prevent $\mathcal{M}_{\text{explorer}}$ from repeating the same error, $\mathcal{M}_{\text{supervisor}}$ will issue an explicit warning reporting the previous failure patterns, guiding $\mathcal{M}_{\text{explorer}}$ toward valid actions and ensuring high-quality trajectory generation.
\subsubsection{Working Memory}
The global working memory $\mathcal{H}$ serves as a fundamental component of MCG, maintaining a record of interaction history that is essential for meaningful exploration. 
At the beginning of each interaction step $t$, the supervisor $\mathcal{M}_{\text{supervisor}}$ compresses the current working memory $\mathcal{H}$ into a concise summary $\tilde{\mathcal{H}}_t$, which is then injected into $\mathcal{M}_{\text{explorer}}$'s prompt to guide the next action.  
At the conclusion of step $t$, $\mathcal{M}_{\text{supervisor}}$ appends a new step record to $\mathcal{H}$ based on the explorer's previous thought $r_t$ and action $a_t$.  
This memory enables the agents to maintain long-horizon consistency and supports rollback decisions.

\begin{algorithm}[h]
  \caption{Supervisor-Guided Step Budget Allocation}
  \label{alg:budget_allocation}
  \begin{algorithmic}
    \STATE {\bfseries Input:} pending application set $\mathcal{A}'_\tau$, total step budget $d_\tau$, allocation weights $\{w_a\}_{a \in \mathcal{A}'_\tau}$
    \STATE {\bfseries Output:} step budgets $\{B_a\}_{a \in \mathcal{A}'_\tau}$

    \IF{$\{w_a\}$ is empty}
        \STATE $\{w_a\}_{a \in \mathcal{A}'_\tau} \gets \frac{1}{|\mathcal{A}'_\tau|} \quad \forall a \in \mathcal{A}'_\tau$
        \hfill // initialize all app weights equally
    \ENDIF

    \STATE $\{B_a\}_{a \in \mathcal{A}'_\tau} \gets 
    \mathcal{M}_{\text{supervisor}}(\mathcal{A}'_\tau,\, d_\tau,\, \{w_a\})$ 
    \hfill // supervisor-based budget allocation

    \STATE \textbf{return} $\{B_a\}_{a \in \mathcal{A}'_\tau}$
  \end{algorithmic}
\end{algorithm}

\begin{algorithm}[h]
  \caption{Adaptive Budget Weight Update}
  \label{alg:weight_update}
  \begin{algorithmic}
    \STATE {\bfseries Input:} current weights $\{w_a\}$, cyclic counter $c_a$, threshold $\gamma$, penalty ratio $p_w$
    \STATE {\bfseries Output:} updated weights $\{w_a'\}$

    \IF{$c_a > \gamma$}
      \STATE $w_a' \gets w_a$
      \STATE $w_a' \gets (1 - p_w) \cdot w_a$ \hfill // penalize current application
      \STATE $w_a' \gets \dfrac{w_a'}{\sum_{a' \in \mathcal{A}'_\tau} w_{a'}'} \quad \forall a \in \mathcal{A}'_\tau$
      \STATE \textbf{return} $\{w_a'\}$
    \ELSE
      \STATE \textbf{return} $\{w_a\}$ \hfill // no update needed
    \ENDIF
  \end{algorithmic}
\end{algorithm}

\clearpage

\subsubsection{Agent Prompts}
\begin{figure*}[h]
\centering
\begin{tcolorbox}[
    colback=myback,        
    colframe=myframe,      
    listing only, 
    title=\textbf{\textit{Prompt for Semantic Difficulty}}
]

\textbf{INTERACTION\_CONTROL\_DIFFICULTY\_PROMPTS}

\textbf{EASY:} 
The target UI element should be clearly visible and associated with an explicit label 
that closely matches the task instruction. 
The target interaction component type is intuitive (e.g., a standard button). 
The agent can directly locate and complete the target interaction 
without requiring complex visual reasoning, semantic alignment, or interaction inference.

\textbf{MEDIUM:} 
The target UI element may exhibit a certain degree of visual or semantic ambiguity. 
The agent is required to perform limited visual reasoning, synonym matching, 
or interact with some complex components (such as list items, dropdown menus, or input fields) 
to correctly map the task instruction to the intended UI action.

\textbf{HARD:} 
The target UI element is difficult to identify directly 
or presented in a non-standard or abstract visual form. 
The interface may contain multiple highly similar candidate elements. 
Moreover, the target components are mostly complex (such as list items, dropdown menus, or input fields). 
The agent must integrate contextual information and perform extensive visual, semantic, 
and interaction reasoning to correctly map the task instruction to executable GUI actions.

\vspace{1em}

\textbf{INSTRUCTION\_UNDERSTANDING\_DIFFICULTY\_PROMPTS}

\textbf{EASY:} 
The task instruction should be expressed as a direct and concrete natural language command. 
It should not involve abstract goals or implicit intentions.

\textbf{MEDIUM:} 
The task instruction may include some abstract goals. 
However, the overall instruction should remain concrete.

\textbf{HARD:} 
The task instruction should be formulated as a high-level goal or a vague intention. 
Successfully completing the task requires the agent to perform non-literal semantic reasoning, 
leveraging common sense, domain knowledge, or user habits to infer missing details 
and fully interpret the intended instruction.

\end{tcolorbox}
\caption{Prompt for semantic difficulty. 
INTERACTION\_CONTROL\_DIFFICULTY\_PROMPTS are used in the exploration stage, where one of the prompt corresponding to the sampled difficulty parameters (easy, medium, or hard) is selected as input to $\mathcal{M}_\text{explorer}$'s prompt.
INSTRUCTION\_UNDERSTANDING\_DIFFICULTY\_PROMPTS are used for inverse instruction synthesis, where one of the prompt corresponding to the sampled difficulty parameters (easy, medium, or hard) is selected as input to $\mathcal{M}_\text{synthesizer}$'s prompt. }
\label{fig:semantic_difficulty_prompt}
\end{figure*}

\begin{figure*}[h]
\centering
\begin{tcolorbox}[
    colback=myback,        
    colframe=myframe,  
    listing only, 
    title=\textbf{\textit{Role-Play Prompt for Explorer}}
]

You are an agent capable of autonomously operating on an Android device.\par
\vspace{0.5em}
Your goal is to generate coherent, realistic, and goal-driven interaction trajectories for training/evaluating GUI agents. Each trajectory should look like a real user trying to accomplish meaningful micro-goals within the allowed apps, while using the available step budget. Avoid obviously meaningless actions, repeated loops, and random tapping. If you finish a micro-goal early, start another related micro-goal, while keeping the overall trajectory semantically coherent.\par
\vspace{0.5em}
- Click/tap on an element on the screen. We have added marks (bounding boxes with numeric indexes on their TOP LEFT corner) to most of the UI elements in the screenshot, use the numeric index to indicate which element you want to click: \\
\hspace*{1em} \{'action\_type': 'click', 'index': $<$target\_index$>$\}. \par
- If the UI element list is missing / unreliable, you may instead click by pixel coordinates (x, y) on the screenshot (in pixels): \\
\hspace*{1em} \{'action\_type': 'click', 'x': $<$x$>$, 'y': $<$y$>$\}. \par
- Long press on an element on the screen, similar with the click action above, use the numeric label on the bounding box to indicate which element you want to long press: \\
\hspace*{1em} \{'action\_type': 'long\_press', 'index': $<$target\_index$>$\}. \par
- Or long press by coordinates: \\
\hspace*{1em} \{'action\_type': 'long\_press', 'x': $<$x$>$, 'y': $<$y$>$\}. \par
- Type text into a text field (this action contains clicking the text field, typing in the text and pressing the enter, so no need to click on the target field to start), use the numeric label on the bounding box to indicate the target text field: \\
\hspace*{1em} \{'action\_type': 'input\_text', 'text': $<$text\_input$>$, 'index': $<$target\_index$>$\} \par
- Or type text by coordinates: \\
\hspace*{1em} \{'action\_type': 'input\_text', 'text': $<$text\_input$>$, 'x': $<$x$>$, 'y': $<$y$>$\} \par
- Open App: \{'action\_type': 'open\_app', 'app\_name': '$<$name$>$'\} (use this to switch between apps) \\
- Press the Enter key: \{'action\_type': 'keyboard\_enter'\} \\
- Navigate to the home screen: \{'action\_type': 'navigate\_home'\} \\
- Navigate back: \{'action\_type': 'navigate\_back'\} \\
- Scroll the screen or a scrollable UI element in one of the four directions, use the same numeric index as above if you want to scroll a specific UI element, leave it empty when scroll the whole screen: \\
\hspace*{1em} \{'action\_type': 'scroll', 'direction': $<$up, down, left, right$>$, 'index': $<$optional\_target\_index$>$\} \par
- Wait for the screen to update: \{'action\_type': 'wait'\}

\end{tcolorbox}
\caption{Role-play prompt for explorer agent $\mathcal{M}_{\text{explorer}}$.}
\label{fig:explorer_role_prompt}
\end{figure*}

\begin{figure*}[h]
\centering
\begin{tcolorbox}[
    colback=myback,        
    colframe=myframe,  
    listing only, 
    title=\textbf{\textit{Action Selection Prompt for Explorer}}
]
\$ROLE\_PLAY\_PROMPT\$\\
Here is a history of what you have done so far: \$HISTORY\_SUMMARY\$\\
Here are the details of the latest steps: \$LATEST\_STEPS\$\\

The current screenshot with bounding boxes and labels added are also given to you.\\
Here is a list of detailed information for some of the UI elements (notice that some elements in this list may not be visible in the current screen and so you can try to scroll the screen to reveal it first), the numeric indexes are consistent with the ones in the labeled screenshot: \$UI\_ELEMENTS\$\\

Now you are in the app: \$CURR\_APP\$\\
Step budget for this app: \$APP\_STEP\_BUDGET\$\\

You need to ensure that your trajectory is semantically coherent and meets the required semantic complexity: \$INTERACTION\_CONTROL\_DIFFICULTY\_PROMPT\$\\
\$ADDITIONAL\_GUIDELINES\$\\

Now output an action from the above list in the correct JSON format, following the reason why you do that. Make sure that the reason explains only the current atomic action, and does not include multiple actions or future plans.\\
Your answer should look like: \\
Reason: ...\\
Action: \{`action\_type': ...\}\\

Your answer:

\end{tcolorbox}
\caption{Action selection prompt for explorer agent $\mathcal{M}_{\text{explorer}}$.}
\label{fig:explorer_action_selection_prompt}
\end{figure*}

\begin{figure*}[h]
\centering
\begin{tcolorbox}[
    colback=myback,        
    colframe=myframe,  
    listing only, 
    title=\textbf{\textit{Step Budget Allocation Prompt for Supervisor}}
]
You are a supervisor planning step budgets for a GUI agent explore trajectory.\par
\vspace{0.5em}
Target apps: \$TARGET\_APPS\$ \par
Total steps: \$TOTAL\_STEPS\$ \par
\vspace{0.5em}
Based on your knowledge of these apps and typical user tasks, allocate an initial step budget to each app. Consider: \\
- App allocation weight: \$WEIGHTS\$ (Apps with higher weights require more steps) \\
- Common sense (e.g., Settings app usually requires more navigation) \par
\vspace{0.5em}
Output your allocation as a JSON object: \{ \\
\hspace*{1em} 'App1': steps1, \\
\hspace*{1em} 'App2': steps2, \\
\hspace*{1em} ... \\
\} \par
\vspace{0.5em}
Ensure the sum equals \$TOTAL\_STEPS\$. \par
\vspace{0.5em}
Your decision:
\end{tcolorbox}
\caption{Step budget allocation prompt for supervisor agent $\mathcal{M}_{\text{supervisor}}$.}
\label{fig:supervisor_budget_prompt}
\end{figure*}

\begin{figure*}[h]
\centering
\begin{tcolorbox}[
    colback=myback,        
    colframe=myframe,  
    listing only, 
    title=\textbf{\textit{Error Detection Prompt for Supervisor}}
]

You are a supervisor monitoring agent behavior for loops and repeated actions. \\
Recent step records: \$WORKING\_MEMORY\$ \\
Current step: \$CURRENT\_STEP\$ \\
Analyze whether the agent is stuck in a loop or repeating meaningless actions. 
If yes, suggest a backtrack point (step number) to retry from a different state. \par
\vspace{0.5em}
Output format: \par
- If error detected: 'backtrack to step $<$N$>$' (where N $<$ \$CURRENT\_STEP\$) \par
- If no issue: 'no backtrack needed' \par
\vspace{0.5em}
Your decision:

\end{tcolorbox}
\caption{Error detection prompt for supervisor agent $\mathcal{M}_{\text{supervisor}}$.}
\label{fig:supervisor_error_prompt}
\end{figure*}

\begin{figure*}[h]
\centering
\begin{tcolorbox}[
    colback=myback,        
    colframe=myframe,  
    listing only, 
    title=\textbf{\textit{Error Warning Prompt for Supervisor}}
]

[WARNING] You were backtracked from step \$STEP\_NUM\$ to step \$BACKTRACK\_STEP\$ due to detected repetitive behavior.\par
\vspace{0.5em}
Previous failed path involved: \$BACKTRACKED\_ACTIONS\$. \par
\vspace{0.5em}
You MUST try a completely different approach: \\
1) Choose a different UI element, \\
2) Use a different action type, \\
3) Navigate to a different part of the app, or \\
4) Use navigate\_back/navigate\_home to reset context.

\end{tcolorbox}
\caption{Error warning prompt for supervisor agent $\mathcal{M}_{\text{supervisor}}$.}
\label{fig:error_warning_prompt}
\end{figure*}

\begin{figure*}[h]
\centering
\begin{tcolorbox}[
    colback=myback,        
    colframe=myframe,  
    listing only, 
    title=\textbf{\textit{History Summarization Prompt for Supervisor}}
]

You are a supervisor summarizing the agent's recent trajectory history.\par
\vspace{0.5em}
Recent step records: \$STEP\_RECORDS\$ \par
Current app: \$CURRENT\_APP\$ \par
\vspace{0.5em}
Generate a concise summary ($\le$ \$MAX\_WORDS\$ words) of what the agent has accomplished so far. Focus on: \\
- Key milestones (app switches, completed micro-tasks) \\
- Current exploration context (what the agent is currently doing) \\
- Important state transitions \par
\vspace{0.5em}
Keep it as a single, brief paragraph to support the next step's decision-making.\par
\vspace{0.5em}
Your summary:
\end{tcolorbox}
\caption{History summarization prompt for supervisor agent $\mathcal{M}_{\text{supervisor}}$.}
\label{fig:supervisor_history_summary_prompt}
\end{figure*}

\begin{figure*}[h]
\centering
\begin{tcolorbox}[
    colback=myback,        
    colframe=myframe,  
    listing only, 
    title=\textbf{\textit{Action Summarization Prompt for Supervisor}}
]

Generate a concise ($\le$ 20 words) description of what the agent did in this step.\par
\vspace{0.5em}
Step number: \$STEP\_NUM\$ \par
Current App: \$CURR\_APP\$ \par
Agent's Reasoning: \$THOUGHTS\$ \par
Action: \$ACTION\$ \par
\vspace{0.5em}
Summarize the action in one sentence. Focus on: \\
- What UI element was interacted with (if mentioned in reasoning) \\
- What operation was performed (click, scroll, input, etc.) \\
- The purpose or goal of this action (inferred from reasoning) \par
\vspace{0.5em}
Your summary:

\end{tcolorbox}
\caption{Action summarization prompt for supervisor agent $\mathcal{M}_{\text{supervisor}}$.}
\label{fig:supervisor_action_summary_prompt}
\end{figure*}

\begin{figure*}[h]
\centering
\begin{tcolorbox}[
    colback=myback,        
    colframe=myframe,  
    listing only, 
    title=\textbf{\textit{Thought Synthesis Prompt for Synthesizer}}
]
You are a Student GUI agent writing training data.\par
\vspace{0.5em}
You are given two Android screenshots: BEFORE (raw screen before executing the action) and AFTER (raw screen after executing the action). \\
Screenshots do NOT contain UI boxes/SoM marks. \par
\vspace{0.5em}
App: \$APP\_NAME\$ \\
Action JSON: \$ACTION\_JSON\$ \\
Target element (if any): \$ELEMENT\_JSON\$ \par
\vspace{0.5em}
\textbf{Task:} \\
Write a short ReAct-style reasoning (thought) that a good GUI agent could have BEFORE taking this action. \\
- The reasoning MUST be consistent with the given action and the inferred instruction. \\
- It should refer to visible UI evidence when possible (e.g., button text/icon), and avoid index numbers. \\
- It must describe ONLY this single step (no multi-step plans). \par
\vspace{0.5em}
\textbf{Return ONLY a valid JSON object with exactly these keys:} \\
\hspace*{1em} \{ \\
\hspace*{2em} "reasoning": "one short first-person thought (why this action now)", \\
\hspace*{2em} "analysis": "brief meta / debug (optional)" \\
\hspace*{1em} \}

\end{tcolorbox}
\caption{Thought synthesis prompt for synthesizer agent $\mathcal{M}_{\text{synthesizer}}$.}
\label{fig:thought_synthesis_prompt}
\end{figure*}

\begin{figure*}[h]
\centering
\begin{tcolorbox}[
    colback=myback,        
    colframe=myframe,  
    listing only, 
    title=\textbf{\textit{Instruction Synthesis Prompt for Synthesizer}}
]
You are a Teacher writing the user's high-level task instruction.\par
\vspace{0.5em}
You are given a step-by-step Android interaction trace (action + student thought). \\
App name requirement: "task\_instruction" MUST explicitly mention the app name(s) exactly as written: \$APP\_HINT\$ \\
Do NOT translate/paraphrase the app name(s). \par
\vspace{0.5em}
Input text requirement: \\
- If the interaction trace contains any input\_text action with a non-empty "text", then "task\_instruction" MUST include that exact text string verbatim. \\
- Required typed texts (verbatim, do NOT modify): \$REQUIRED\_INPUT\_TEXTS\$ \par
\vspace{0.5em}
Instruction semantic complexity requirement: \$INSTRUCTION\_DIFFICULTY\_PROMPT\$ \par
\vspace{0.5em}
Interaction trace:  \$STEPS\$ \par
\vspace{0.5em}
Task: \\
- Write ONE task-level instruction (a single sentence) that captures the user's overall goal. \\
- The instruction MUST sound like a natural human request to an assistant. \\
- Follow the semantic complexity requirement strictly. \par
\vspace{0.5em}
Return ONLY a valid JSON object with exactly these keys: \\
\{ \\
\hspace*{1em} "task\_instruction": "one sentence", \\
\hspace*{1em} "analysis": "brief reasoning" \\
\}
\end{tcolorbox}
\caption{Instruction synthesis prompt for synthesizer agent $\mathcal{M}_{\text{synthesizer}}$.}
\label{fig:instruction_synthesis_prompt}
\end{figure*}

\begin{figure*}[h]
\centering
\begin{tcolorbox}[
    colback=myback,        
    colframe=myframe,  
    listing only, 
    title=\textbf{\textit{Role-Play Prompt for Student}}
]

You are an Android GUI agent. Your goal is to complete tasks given a high-level instruction, action history, and current screenshot with its UI tree.\par
\vspace{0.5em}
\textbf{STRICT OUTPUT RULES:}\par
1. You MUST ONLY output a single JSON object per step. \\
2. The "action\_type" value MUST be chosen EXCLUSIVELY from the list below. Do NOT use any other synonyms or any variations. \\
3. NO conversational text, explanations, or additional keys are allowed outside the JSON.\par
\vspace{0.5em}
\textbf{ALLOWED ACTION TYPES:}\par
- Click (preferred with SoM marks): \{'action\_type': 'click', 'index': $<$target\_index$>$\} \\
- Click (fallback by pixel coords): \{'action\_type': 'click', 'coordinates': [x, y]\} \\
- Long Press (preferred with SoM marks): \{'action\_type': 'long\_press', 'index': $<$target\_index$>$\} \\
- Long Press (fallback by pixel coords): \{'action\_type': 'long\_press', 'coordinates': [x, y]\} \\
- Type Text (preferred with SoM marks): \{'action\_type': 'input\_text', 'text': '$<$text$>$', 'index': $<$target\_index$>$\} \\
- Type Text (fallback by pixel coords): \{'action\_type': 'input\_text', 'text': '$<$text$>$', 'coordinates': [x, y]\} \\
- Press Enter: \{'action\_type': 'keyboard\_enter'\} \\
- Home: \{'action\_type': 'navigate\_home'\} \\
- Back: \{'action\_type': 'navigate\_back'\} \\
- Scroll: \{'action\_type': 'scroll', 'direction': '$<$up|down|left|right$>$', 'coordinates': [x, y]\} (Use null coordinates for full-screen) \\
- Open App: \{'action\_type': 'open\_app', 'app\_name': '$<$name$>$'\} (!MANDATORY for starting apps!) \\
- Wait: \{'action\_type': 'wait'\} (Use when the screen is not ready)\par
\vspace{0.5em}
\textbf{CONSTRAINT CHECK:}\par
Before outputting, verify that your "action\_type" matches one of the strings above exactly. If it is not "click", "long\_press", "input\_text", "keyboard\_enter", "navigate\_home", "navigate\_back", "scroll", "open\_app", or "wait", it is FORBIDDEN. If using coordinates, point to the center of the target element. If using index, make sure it matches the marked UI element.

\end{tcolorbox}
\caption{Role-play prompt for student agent $\mathcal{M}_{\text{student}}$.}
\label{fig:role_play_student_prompt}
\end{figure*}

\begin{figure*}[h]
\centering
\begin{tcolorbox}[
    colback=myback,        
    colframe=myframe,  
    listing only, 
    title=\textbf{\textit{Action Selection Prompt for Student}}
]
\$STUDENT\_ROLE\_PLAY\_PROMPT\_TEMPLATE\$ \par
\vspace{0.5em}
\textbf{Examples} \par
\vspace{0.5em}
Example 1 \\
Task instruction: Open Chrome and visit 'github.com'. \\
Latest steps: \\
Reason: To browse the website, I first need to launch the Chrome browser. I will use the 'open\_app' action. \\
Action: \{'action\_type': 'open\_app', 'app\_name': 'Chrome'\} \par
\vspace{0.5em}
Example 2 \\
Task instruction: Create a new note in Markor. \\
Latest steps: Before Step1: To create new note in Markor, I need to start the Markor app using the 'open\_app' action., action: \{'action\_type': 'open\_app', 'app\_name': 'Markor'\} \\
Reason: I have opened Markor. Now I need to click the 'Add' button at [900, 2100] to create a new markdown file. \\
Action: \{'action\_type': 'click', 'coordinates': [900, 2100]\} \par
\vspace{0.5em}
Example 3 \\
Task instruction: Scroll to see more tracks in Retro Music. \\
Latest steps: Before Step1: To see the music tracks, I first need to open the Retro Music app., action: \{'action\_type': 'open\_app', 'app\_name': 'Retro Music'\} \\
Reason: I am looking for a specific song. I will scroll down to reveal more items in the music list. \\
Action: \{'action\_type': 'scroll', 'direction': 'down', 'coordinates': null\} \par
\vspace{0.5em}
\textbf{Current Task} \par
\vspace{0.5em}
Here is your action history:  \$HISTORY\$ \\
Task instruction: \$TRAJECTORY\_INSTRUCTION\$ \\
Latest steps: \$LATEST\_STEPS\$ \\
Accessibility tree: \$UI\_ELEMENTS\$ \par
\vspace{0.5em}
Please generate your thoughts and action for the next step. Make sure that the reason explains only the current atomic action, and does not include multiple actions or future plans. Your answer should look like: \par
\vspace{0.5em}
Reason: [Brief thoughts] \\
Action: \{'action\_type': '...'\} \par
\vspace{0.5em}
Your Answer:
\end{tcolorbox}
\caption{Action selection prompt for student agent $\mathcal{M}_{\text{student}}$.}
\label{fig:student_action_selection_prompt}
\end{figure*}

\begin{figure*}[h]
\centering
\begin{tcolorbox}[
    colback=myback,        
    colframe=myframe,  
    listing only, 
    title=\textbf{\textit{Action Summarization Prompt for Student}}
]

You are a GUI agent maintaining a working memory by yourself.\par
\vspace{0.5em}
Recent step records: \$STEP\_RECORDS\$ \par
Current app: \$CURRENT\_APP\$ \par
\vspace{0.5em}
Generate a concise summary ($\le$ \$MAX\_WORDS\$ words) of what has been accomplished so far. Focus on: \\
- Key actions taken (app switches, button clicks, text inputs) \\
- Current progress towards the task goal \\
- Important state changes \par
\vspace{0.5em}
Keep it as a single, brief paragraph to support the next action decision.\par
\vspace{0.5em}
Your summary:

\end{tcolorbox}
\caption{Action Summarization prompt for student agent $\mathcal{M}_{\text{student}}$.}
\label{fig:student_action_summarization_prompt}
\end{figure*}

\end{document}